\documentclass[]{imag-ms-template}

\title{Deep Riemannian Networks for \textcolor{black}{End-to-End} EEG Decoding}

\author{Daniel Wilson${}^{\ast1, 2}$, Robin T. Schirrmeister${}^{1, 2}$, Lukas A. W. Gemein${}^1$, Tonio Ball${}^{1, 2}$\\
{\small $^{1}$Neuromedical A.I. Lab, Department of Neurosurgery, Medical Center - University of Freiburg,}\\
{\small Faculty of Medicine, University of Freiburg, Freiburg, Germany}\\
{\small $^{2}$BrainLinks-BrainTools, IMBIT (Institute for Machine-Brain Interfacing Technology),} \\ {\small University of Freiburg, 79110 Freiburg im Breisgau, Germany}\\
{\small $^\ast$Correspondence:  daniel.wilson@uniklinik-freiburg.de}
}

\begin{document} 

\maketitle

\begin{acronym}
\acro{eeg}[EEG]{Electroencephalography}
\acro{meg}[MEG]{Magnetoencephalography}
\acro{fmri}[fMRI]{functional Magnetic Resonance Imaging}
\acro{bci}[BCI]{Brain-Computer Interface}
\acro{dti}[DTI]{Diffusion Tensor Imaging}
\acro{ecg}[ECG]{Electrocardiography}

\acro{spd}[SPD]{symmetric-positive definite}
\acro{psd}[PSD]{positive semi-definite}
\acro{pd}[PD]{positive-definite}

\acro{dl}[DL]{Deep-Learning}
\acro{sl}[SL]{Shallow-Learning}
\acro{rbd}[RBD]{Riemannian-Geometry-based decoder}
\acro{ml}[ML]{Machine Learning}
\acro{cv}[CV]{cross-validation}

\acro{bo}[BO]{Bayesian Optimiser}
\acro{fbopt}[FB-Opt]{Filterbank Optimisation}
\acro{eespdnet}[EE(G)-SPDNet]{end-to-end EEG SPDNet}
\acro{fbspdnet}[FBSPDNet]{Filterbank Optimised EEG SPDNet}
\acro{lbl}[LBL]{layer-by-layer}

\acro{chind}[ChInd]{channel independent filtering}
\acro{chspec}[ChSpec]{channel specific filtering}

\acro{dnn}[DNN]{Deep Neural-Network}
\acro{drn}[DRN]{Deep Riemannian Network}
\acro{svm}[SVM]{Support-Vector Machine}
\acro{rsvm}[rSVM]{Riemannian Support-Vector Machine}
\acro{mdm}[MDM]{Minimum Distance to Mean Classifier}
\acro{rmdm}[rMDM]{Minimum Distance to Riemannian-Mean Classifier}
\acro{csp}[CSP]{Common Spatial Patterns}
\acro{fbcsp}[FBCSP]{Filter Bank Common Spatial Patterns}
\acro{mlp}[MLP]{Multi-Layer Perceptron}


\acro{rbf}[RBF]{Radial-Basis Function}
\acro{oas}[OAS]{Oracle-Approximating Shrinkage}
\acro{lwf}[LWF]{Ledoit-Wolf}
\acro{scm}[SCM]{Sample Covariance Matrix}

\acro{logeuc}[LEM]{Log-Euclidean Metric}
\acro{airm}[AIRM]{Affine Invariant Riemannian Metric}


\acro{relu}[ReLU]{Rectified Linear Unit}

\acro{hgd}[HGD]{High-Gamma Dataset}
\end{acronym}


\begin{abstract}
State-of-the-art performance in electroencephalography (EEG) decoding tasks is currently often achieved with either Deep-Learning (DL) or Riemannian-Geometry-based decoders (RBDs). 
Recently, there is growing interest in Deep Riemannian Networks (DRNs) possibly combining the advantages of both previous classes of methods.
However, there are still a range of  topics where additional insight is needed to pave the way for a more widespread application of DRNs in EEG.
These include architecture design questions such as network size and end-to-end ability.
How these factors affect model performance has not been explored.
Additionally, it is not clear how the data within these networks is transformed, and whether this would correlate with traditional EEG decoding. 
Our study aims to lay the groundwork in the area of these topics through the analysis of DRNs for EEG with a wide range of hyperparameters.
Networks were tested on \textcolor{black}{five} public EEG datasets and compared with state-of-the-art ConvNets.

Here we propose \ac{eespdnet}, and we show that this wide, end-to-end DRN can outperform the ConvNets, and in doing so use physiologically plausible frequency regions.
We also show that the end-to-end approach learns more complex filters than traditional band-pass filters targeting the classical alpha, beta, and gamma frequency bands of the EEG, and that performance can benefit from channel specific filtering approaches. 
Additionally, architectural analysis revealed areas for further improvement due to the possible \textcolor{black}{under utilisation} of Riemannian specific information throughout the network.
Our study thus shows how to design and train DRNs to infer task-related information from the raw EEG without the need of handcrafted filterbanks and highlights the potential of end-to-end DRNs such as \ac{eespdnet} for high-performance EEG decoding.
\end{abstract}

\section{Introduction}

Optimizing the amount of information that can be extracted from brain signals such as the \ac{eeg} is crucial for \ac{bci} performance and thus for the development of viable \ac{bci} applications. 
The multivariate \ac{eeg} data is inherently complex as task-relevant information can be found \textcolor{black}{at} multiple frequencies, as well as in the correlation structure of the signals across different electrodes.
Also, the generative processes which give rise to the \ac{eeg} signals are not fully understood. 
To handle such complex data, most current \ac{eeg}-\acp{bci} use \ac{ml} for the analysis of the recorded brain signals.

Currently, the state-of-the-art in \ac{eeg}-\ac{bci} decoding with \ac{ml} is often achieved through two relatively distinct strategies that have been developed in parallel over the last decade, \acf{dl} \citep{lecunDeepLearning2015, schirrmeisterDeepLearningConvolutional2017, lawhernEEGNetCompactConvolutional2018} and \acfp{rbd}  \citep{barachantClassificationCovarianceMatrices2013, congedoRiemannianGeometryEEGbased2017, ygerRiemannianApproachesBrainComputer2017, lotteReviewClassificationAlgorithms2018, chevallierRiemannianGeometryCombining2022}.
\acp{rbd} employ concepts from Riemannian geometry to leverage inherent geometrical properties of the covariance matrix of the \ac{eeg} \citep{barachantClassificationCovarianceMatrices2013, congedoRiemannianGeometryEEGbased2017, ygerRiemannianApproachesBrainComputer2017, lotteReviewClassificationAlgorithms2018}.
The covariance matrix captures task relevant \textcolor{black}{spatial} information in not only the variances of the signals from single electrodes, but also the covariance between electrode pairs.
\ac{dl} uses multi-layered artificial neural network models and training using backpropagation to learn to extract hierarchical structure from input data, which has led to state-of-the-art performance in a number of fields, including \ac{eeg} decoding \citep{lecunDeepLearning2015, schirrmeisterDeepLearningConvolutional2017, lawhernEEGNetCompactConvolutional2018}.
The network's multilayered structure means that they are able to learn sequential and/or parallel data transformations which were not pre-defined manually.
This has led to the development of \textit{end-to-end} models for \ac{eeg} decoding (such as those by \citet{schirrmeisterDeepLearningConvolutional2017}) that can be fed raw or minimally preprocessed data.
An advantage of such end-to-end analyses is that processing steps such as data whitening or feature extraction are implicitly learned and optimised jointly with the classification during model training.
Therefore, while \ac{dl} can be viewed as a more general classification model, in principle suitable to extract and use arbitrary learned \ac{eeg} features, \acp{rbd} are specialised to optimally leverage the information contained in the \ac{eeg} covariance structure.

In the last few years, the success of \acp{rbd} and \ac{dl} on EEG decoding and on other domains has motivated a number of works that sought to combine these two methods, leading to various types of \acfp{drn}.
Some are orientated towards manifold valued data \citep{chakrabortyManifoldNetDeepNetwork2018, chakrabortyManifoldNormExtendingNormalizations2020, pennecManifoldvaluedImageProcessing2020}, such as the arrays produced by \ac{dti} but not \ac{eeg}.
Others are designed to operate on \ac{spd} matrices (the category of matrix to which the covariance matrix belongs) \citep{huangRiemannianNetworkSPD2016, dongDeepManifoldLearning2017, liuLearningNeuralBagofMatrixSummarization2019, acharyaCovariancePoolingFacial2018, yuSecondorderConvolutionalNeural2017, brooksRiemannianBatchNormalization2019, suhRiemannianEmbeddingBanks2021, sukthankerNeuralArchitectureSearch2021}.
While some of these methods are demonstrated on facial/image recognition datasets \citep{huangRiemannianNetworkSPD2016, dongDeepManifoldLearning2017, brooksRiemannianBatchNormalization2019, acharyaCovariancePoolingFacial2018, yuSecondorderConvolutionalNeural2017, liuLearningNeuralBagofMatrixSummarization2019}, some have also been demonstrated with \ac{eeg} \citep{liuLearningNeuralBagofMatrixSummarization2019, suhRiemannianEmbeddingBanks2021, juTensorCSPNetNovelGeometric2022a, majidovEfficientClassificationMotor2019, yangMLPRiemannianCovariance2020, koblerSPDDomainspecificBatch2022, zhangRFNetRiemannianFusion2020}.
One particular model of note is the SPDNet \citep{huangRiemannianNetworkSPD2016} (demonstrated on action, emotion and facial recognition data) which was designed to mimic some of the functions of a convolutional network, but entirely on \ac{spd} data.
Table \ref{Tab: DeepRieEEG_FB} summarises the literature concerning Deep Riemannian Networks for \ac{eeg}.
\textcolor{black}{
Of the literature seen in Table \ref{Tab: DeepRieEEG_FB}, SPDNet-like networks are the most widely used \acp{drn}, with several of the cited works using various combinations of its layers within their architecture \citep{liuLearningNeuralBagofMatrixSummarization2019, suhRiemannianEmbeddingBanks2021, hajinorooziPredictionFatiguerelatedDriver2017, juTensorCSPNetNovelGeometric2022a, koblerSPDDomainspecificBatch2022}.
Since these networks excel at extracting relevant spatial-spectral information from the \ac{eeg}, the majority of these models \citep{yangMLPRiemannianCovariance2020, suhRiemannianEmbeddingBanks2021, juTensorCSPNetNovelGeometric2022a, koblerSPDDomainspecificBatch2022, zhangRFNetRiemannianFusion2020, liuLearningNeuralBagofMatrixSummarization2019, majidovEfficientClassificationMotor2019} were demonstrated on motor EEG data - where the spectral power at different electrodes is known to be a prominent source of information.
The differing architectures for these works suggests there to be no clear choice for SPDNet architecture on \ac{eeg}, including the filterbanking applied pre-SPDNet.
}

Since a variety of subject-specific frequency bands may contain task-relevant information, \ac{eeg} decoding pipelines regularly feature a filterbanking step designed to separate the incoming signal into multiple frequency bands \citep{schirrmeisterDeepLearningConvolutional2017, angFilterBankCommon2012, islamMultibandTangentSpace2018, akterMultibandEntropybasedFeatureextraction2020, yangMLPRiemannianCovariance2020, majidovEfficientClassificationMotor2019, juTensorCSPNetNovelGeometric2022a, koblerSPDDomainspecificBatch2022, thomasAdaptiveFilterBank2008, thomasDiscriminativeFilterBankSelection2009, wuNewSubjectSpecificDiscriminative2021,zhangRFNetRiemannianFusion2020,belwafiAdaptiveEEGFiltering2014,belwafiEmbeddedImplementationBased2018, mousaviTemporallyAdaptiveCommon2019, maneFBCNetMultiviewConvolutional2021}.
For methods like \acp{rbd} that use variance information this filterbanking approach is a logical step: Different frequency bands are well-known to reflect different aspects of the brain's functional connectivity, and hence the covariance structure of the recorded \ac{eeg} signals is strongly affected by the frequency composition of the input signal.
While filtering or filterbanking is a ubiquitous step for \ac{eeg} decoding, the literature summarised in Table \ref{Tab: DeepRieEEG_FB} show  that for \acp{drn} there is no clear-cut filterbanking method or architecture.
\textcolor{black}{The} filterbank can \textcolor{black}{either be constructed from a set of predefined candidate} filters or by freely learning the filters (e.g. learning the cut-offs, or convolutional kernel).
We will call the first construction method pre-defined filterbanks and the second method learnable filterbanks.

Pre-defined filterbanks have been used in many related areas, including \ac{csp} \citep{angFilterBankCommon2012, thomasAdaptiveFilterBank2008, thomasDiscriminativeFilterBankSelection2009, belwafiEmbeddedImplementationBased2018}, \acp{rbd} \citep{islamMultibandTangentSpace2018, wuNewSubjectSpecificDiscriminative2021} and \acp{drn} \citep{juTensorCSPNetNovelGeometric2022a, zhangRFNetRiemannianFusion2020, yangMLPRiemannianCovariance2020} and consist of an array of specific filters.
Despite being explicitly defined, these filterbanks are frequently large and the filters are often searched through in order to reduce feature dimension \citep{thomasAdaptiveFilterBank2008, thomasDiscriminativeFilterBankSelection2009, angFilterBankCommon2012, belwafiEmbeddedImplementationBased2018, islamMultibandTangentSpace2018, wuNewSubjectSpecificDiscriminative2021}, although some models use all filters and do not explicitly reduce feature dimension at this stage \citep{zhangRFNetRiemannianFusion2020, maneFBCNetMultiviewConvolutional2021, juTensorCSPNetNovelGeometric2022a, yangMLPRiemannianCovariance2020}.
Search metrics vary, but all are intended as a proxy for class separability, be it the Fisher information ratio \citep{thomasAdaptiveFilterBank2008, thomasDiscriminativeFilterBankSelection2009}, mutual information \citep{islamMultibandTangentSpace2018} or a statistical test \citep{wuNewSubjectSpecificDiscriminative2021}.
Due to their explicit nature, they are easy to interpret and implement, but also require \textit{hand-crafting} which may preclude some level of domain knowledge.
\textcolor{black}{Making the filterbank learnable eliminates this limitation and enables true end-to-end model training.}

Learnable filterbanks are those that are flexible and not limited to a pre-defined set of filters but instead search a much larger space as part of an optimisation process, and they have been generally explored within the confines of \ac{dl} \citep{schirrmeisterDeepLearningConvolutional2017, koblerSPDDomainspecificBatch2022, liChannelProjectionMixedScaleConvolutional2019, mousaviTemporallyAdaptiveCommon2019}.
These deep models have multiple parallel convolutional layers which are intended to mimic the effects of band-pass filtering on \ac{eeg}, first introduced by \citet{schirrmeisterDeepLearningConvolutional2017}.
However, the convolutional layer is not guaranteed to learn a strict bandpass filter, through random initialisation and the following optimisation it may also learn a notch, low-pass, high-pass filter or even a multi-band filter.
The process of optimising the convolutional layer, is essentially searching the band-space with the backpropagated loss of the network serving as the search metric, which, it can be argued, is a more direct measurement of class separability.
This search process is not limited to a pre-defined set of bands/band-types and is inherently less granular and also more adaptable to datasets with a wider range of frequencies (such as those designed to measure frequencies in the high-gamma region).
Learnable filterbanks remain underutilised in \ac{eeg} decoding, particularly in a \acp{drn}.

As can be seen from Table \ref{Tab: DeepRieEEG_FB} there is a relatively small amount of work exploring \acp{drn} in the context of \ac{eeg} \citep{hajinorooziDriverFatiguePrediction2017, liuLearningNeuralBagofMatrixSummarization2019, majidovEfficientClassificationMotor2019, yangMLPRiemannianCovariance2020, suhRiemannianEmbeddingBanks2021, juTensorCSPNetNovelGeometric2022a, koblerSPDDomainspecificBatch2022, zhangRFNetRiemannianFusion2020}.
Furthermore a number of these models have Riemannian transforms prepended onto standard deep networks \citep{majidovEfficientClassificationMotor2019, yangMLPRiemannianCovariance2020, zhangRFNetRiemannianFusion2020}, as opposed to being based on the SPDNet \citep{huangRiemannianNetworkSPD2016}, where the data is manipulated in the Riemannian space through many layers of the network \citep{hajinorooziDriverFatiguePrediction2017, liuLearningNeuralBagofMatrixSummarization2019, suhRiemannianEmbeddingBanks2021, juTensorCSPNetNovelGeometric2022a, koblerSPDDomainspecificBatch2022}.
Moreover, even fewer of these methods used a multispectral approach \citep{majidovEfficientClassificationMotor2019, yangMLPRiemannianCovariance2020, zhangRFNetRiemannianFusion2020, juTensorCSPNetNovelGeometric2022a, koblerSPDDomainspecificBatch2022} and most used a pre-defined filterbank \citep{majidovEfficientClassificationMotor2019, yangMLPRiemannianCovariance2020, zhangRFNetRiemannianFusion2020, juTensorCSPNetNovelGeometric2022a} with the exception of the domain adaptation focused TSMNet \citep{koblerSPDDomainspecificBatch2022} \textcolor{black}{(see Table \ref{Tab: DeepRieEEG_FB})}.
This is despite the theoretical advantages that learnable filterbanks possess.

Therefore, how \acp{drn} can be best equipped with learnable filterbanks \textcolor{black}{and how to gain neurophysiological insights from the learned filterbanks} remains unclear.
The learning of the filterbank can be achieved in an end-to-end manner, with parallel convolutions, or with a black box optimiser using typical bandpass filters, \textcolor{black}{and it is not known} which method is superior.
In addition to this, the size of the filterbank and its channel specificity can also be altered.
Furthermore, how much \acp{drn} use well known regions in the \ac{eeg} (e.g. alpha, beta and high-gamma for motor movement) has gained only limited insight \citep{koblerSPDDomainspecificBatch2022, juTensorCSPNetNovelGeometric2022a}.
It is not yet clear how these factors will affect performance, or whether \acfp{drn} with learnable filterbanks can compete with state-of-the-art networks.

\begin{table}
\begin{threeparttable}
\caption{\color{black}
\textbf{Deep Riemannian Networks for EEG}
Table showing previous \acp{drn} for EEG along with filterbank architectures as they would be described in this study.
For comparison our best performing proposed model is also included.
All models, with the exception of the work by \citet{hajinorooziDriverFatiguePrediction2017}, were evaluated on motor data.
$N_f$ denotes the number of filters in the filterbank (for channel independent filters) or the number of filters per electrode (for channel specific filters).
\label{Tab: DeepRieEEG_FB}}
{\color{black}\begin{tabular}{@{}llll}
\toprule
     Authors & $N_f$ & Filter Specificity & Learnable or Pre-defined \\
\midrule
     \citet{hajinorooziDriverFatiguePrediction2017} & 1 & Channel Independent & Pre-defined \\
     \citet{liuLearningNeuralBagofMatrixSummarization2019} & 1 & Channel Independent & Pre-defined\\
     \citet{majidovEfficientClassificationMotor2019} & 4 & Channel Independent & Pre-defined\\
     \citet{yangMLPRiemannianCovariance2020} & 43 & Channel Independent & Pre-defined\\
     \citet{zhangRFNetRiemannianFusion2020} & 50 & Channel Independent & Pre-defined\\
     \citet{suhRiemannianEmbeddingBanks2021} & 1 & Channel Independent & Pre-defined\\
     \citet{juTensorCSPNetNovelGeometric2022a} & 9 & Channel Independent & Pre-defined\\
     \citet{koblerSPDDomainspecificBatch2022} & 4 & Channel Independent & Learnable \\
     \textbf{EEGSPDNet} (proposed) & 8 & Channel Specific & Learnable\\
\bottomrule
\end{tabular}}
\end{threeparttable}
\end{table}

Therefore, we designed a study to address these questions, wherein we propose and systematically compare two SPDNet-based \acp{drn} with learnable filterbanks for \ac{eeg}, the \acf{eespdnet}\textcolor{black}{,} and the \textcolor{black}{\ac{fbspdnet}}.
\ac{eespdnet} is a second-order convolutional neural network \citep{brooksSecondOrderNetworksPyTorch2019, yuSecondorderConvolutionalNeural2017, liSecondOrderConvolutionalNeural2020}, comprised of a convolutional layer designed to mimic filterbanking, followed by a covariance pooling layer and then an SPDNet.
\ac{fbspdnet} uses a \textcolor{black}{black-box} optimiser to search \textcolor{black}{for an} optimal array of bandpass filters.
The post-filterbank data is then used to generate the covariance matrices that are passed to the SPDNet.
\textcolor{black}{These two models embody distinct methodologies for deep-\acp{rbd} utilizing learned filterbanks. 
They are systematically evaluated and contrasted across various parameters, including filter type, filterbank channel specificity, and filterbank size, among others.}

The design of the \ac{eespdnet} allows for the learning of a custom filterbank from the input data while also being able to use Riemannian geometry to process data on the \ac{spd} manifold.
The \ac{eespdnet} can be separated into two stages, a \textcolor{black}{convolutional stage followed by a} standard SPDNet.
The SPDNet stage reduces the size of the input \ac{spd} matrix, layer-by-layer, while preserving class discriminability before applying a Riemannian transform to whiten the data prior to classification.
The convolutional layer filters the \ac{eeg} signals and, over multiple iterations, becomes optimised to select the ideal frequencies for the covariance matrix.
This removes the need for explicit filterbank architecture design, offering a more precise search of the frequency space.
This precision can be enhanced as the convolutional filters can be optimised for each individual \textcolor{black}{input channel}, creating \textcolor{black}{a \ac{chspec}} filterbank, something that could be cumbersome to design manually.
Furthermore, the convolutional layer allows \ac{eespdnet} to be flexible in adapting to datasets of different sampling rates (and therefore different frequency ranges) since the filterbank architecture does not have to be redesigned, nor the data low-pass filtered to accommodate the filterbank.
Overall, the convolutional layer prepended to the SPDNet allows it to learn an optimal filterbank, as well as potentially create a filterbank composed of various filter types (notch, bandpass etc).

The \ac{fbspdnet} uses regular bandpass filters in conjunction with a \acf{bo} to search the filterbank space, as opposed to the convolution in \ac{eespdnet} - the SPDNet stage remains the same.
At each step in the iterative search, a filterbank is constructed which is used to filter the data before creating the set of \ac{spd} matrices.
The performance of the filterbank is evaluated via cross-validation with a \textcolor{black}{lightweight proxy classifier} on the aforementioned matrices.
The matrices created from the highest scoring filterbank are then passed to the SPDNet.
The \ac{fbspdnet} possesses almost all the previously mentioned advantages of the learnable filterbank approach such as \textcolor{black}{channel} specific architecture and adaptability between datasets, it is restricted only in terms of its filter type.

The bandpass filters used by \ac{fbspdnet} are more rigid in their approach to filtering when compared to the convolutional kernel, which may learn things other than a bandpass filter, but this also gives the \ac{fbspdnet} a smaller space to search.
On the whole, \ac{fbspdnet} represents a model that is somewhere in-between the \ac{eespdnet} and the traditional filterbank approach, with the main difference to the traditional methods being that its array of bandpass filters are learned, rather than explicitly pre-defined.
\textcolor{black}{
Additionally, we also test an \ac{eespdnet} variant, where the initial convolutional layers have been swapped for Sinc layers, adapted from SincNet \citep{ravanelliInterpretableConvolutionalFilters2019}.
This model variant helps bridge the gap between \ac{eespdnet} and \ac{fbspdnet}, as it uses the same bandpass filtering as \ac{fbspdnet}, but the filterbank is optimised jointly with the rest of the network, as in \ac{eespdnet}.
}
Through the experiments performed in this study, we show that our models are capable of outperforming other state-of-the-art classifiers, \textcolor{black}{namely the convolutional networks ShallowFBCSP, Deep4 \citep{schirrmeisterDeepLearningConvolutional2017}, EEGNetv4 \citep{lawhernEEGNetCompactConvolutional2018} and the concurrently developed \ac{drn} TSMNet \citep{koblerSPDDomainspecificBatch2022} with statistically significant improvements achieved by \ac{eespdnet} against every model except TSMNet.}
\textcolor{black}{Broadly these results also demonstrate the effectiveness of \acp{drn} (\ac{eespdnet}, \ac{fbspdnet} and TSMNet) against traditional convolutional networks (Deep4Net, ShallowFBCSPNet and EEGNetv4).}
These improvements were shown on \textcolor{black}{5} public \ac{eeg} \textcolor{black}{motor} datasets: Schirrmeister2017 \citep{schirrmeisterDeepLearningConvolutional2017}, BNCI2014001 \textcolor{black}{\citep{leebBrainComputerCommunication2007}, BNCI2014004 \citep{tangermannReviewBCICompetition2012}, Lee2019 MI \citep{leeEEGDatasetOpenBMI2019} and Shin2017A \citep{shinOpenAccessDataset2017}.}

In addition to analysing the performance of the different models, we also analysed some of the features learnt by the networks, and investigated the network's layer-by-layer performance.
We identified the frequencies that contribute most to classifier performance by looking at the frequency gain caused by the trained filterbanks.
We also visualised and tested separability of the data as it passes through the network in both Euclidean and Riemannian spaces.


\section{Methods}

\subsection{Outline}

The structure of our paper is as follows.
The following methods section will cover: the necessary theoretical background (Section \ref{Sec: Rie Overview}), details of \ac{eespdnet} \& \ac{fbspdnet} (Sections \ref{Sec: EESPDNet Methods} \& \ref{Sec: fbspdnet Methods}), \textcolor{black}{network design choices common to both proposed models (Section \ref{Sec: Methods/NetworkDesignChoices}} specific \ac{spd} matrix operations (\ref{Sec: SPD matrix ops}), methods used for analysis (Section \ref{Sec: Methods/Analyses}), the datasets that were used (Section \ref{Sec: Methods/Datasets}), \textcolor{black}{the experimental procedure (Section \ref{Sec: Methods/Procedure} and the statistical methods used for presenting certain results Section \ref{Sec: Methods/Statistics}}.
After this the results will be presented with figures, in a series of key findings (Section \ref{Results})\textcolor{black}{,} before a discussion of the main points (Section \ref{Discussion}) and the subsequent conclusion (Section \ref{Conclusion}).

\subsection{Riemannian Methods Overview}
\label{Sec: Rie Overview}

Riemannian methods for \ac{eeg} are usually applied on the sample covariance matrix\footnote{The sample covariance matrix is the most common, but any other \ac{spd} matrix would be possible.} of the \ac{eeg} electrodes:
\begin{equation}
    C = \frac{1}{N_t - 1}\textcolor{black}{TT}^\top
\label{Eq:SCM}
\end{equation}
where $\textcolor{black}{T} \in \mathbb{R}^{N_e \times N_t}$.
For \ac{eeg}, $T$ would represent a single trial of $N_e$ electrodes and $N_t$ samples.
\textcolor{black}{Equation \ref{Eq:SCM} is the equation for the \ac{scm}, which is, by construction, both symmetric and positive semi-definite (all its eigenvalues are greater than or equal to zero).
Furthermore, if sufficient data is used ($N_t$ > $N_e$) and $\textcolor{black}{T}$ has full-rank (i.e. no interpolated channels), then $C$ will be positive definite (all eigenvalues are greater than zero).}

\textcolor{black}{\acp{rbd} exploit geometric properties of the \ac{spd} space, in which the computed \acp{scm} reside.
They} use tools from Riemannian geometry to make computations on the \ac{spd} manifold, which has helped them to achieve state-of-the-art performance.
\textcolor{black}{These computations rely on the ability to accurately measure distances on the curved, smooth SPD manifold, properties that arise from the non-linear but open condition of positive-definiteness.
When equipped with an appropriate metric (which defines a smoothly varying inner product on each tangent space of the manifold), the SPD manifold becomes a Riemannian manifold.}
\textcolor{black}{Straight lines or \textit{geodesics}, represent the shortest distances between points on the manifold, however they may differ depending on the chosen metric.}
In practice, the \ac{spd} matrices are mapped to the Euclidean \textcolor{black}{tangent} space using Riemannian geometry, \textcolor{black}{preserving} their structure \textcolor{black}{from} the \textcolor{black}{\ac{spd}} manifold (this is how the \ac{rsvm} is implemented here, a Riemannian transform followed by SVM classification).
This mapping can be \textcolor{black}{understood} as a form of data whitening being applied to the \ac{spd} matrices.
\textcolor{black}{For further information on the mathematics and details of Riemannian Geometry and its applications to EEG, we direct the reader to \citet{barachantClassificationCovarianceMatrices2013} \& \citet{ygerRiemannianApproachesBrainComputer2017} and for some specific details on Riemannian metrics we suggest \citet{huangLogEuclideanMetricLearning2015} \& \citet{arsignyLogEuclideanMetricsFast2006}.}

This study employs the SPDNet, which was first introduced by \citet{huangRiemannianNetworkSPD2016}, and is a deep network designed to "\textit{non-linearly learn desirable \ac{spd} matrices on Riemannian manifolds}" \citep{huangRiemannianNetworkSPD2016}.
The SPDNet uses three \ac{spd} specific layers to process the input data.
BiMap layers are used to reduce \ac{spd} feature dimension and enhance discriminability.
For the $k^{th}$ layer this is done to an input matrix $\textcolor{black}{C}_k$ as follows:
\begin{equation}
    \textcolor{black}{C}_{k} = f_b^{(k)}(\textcolor{black}{C}_{k-1}, W_{k}) = W_{k}^{\top} \textcolor{black}{C}_{k-1} W_{k}
    \label{Eq:BiMap}
\end{equation}
where $W_k$ is the learned weight matrix (must be constrained to a Stiefel manifold of dimension $d_{k-1} \times d_{k}$).

The ReEig layers aim to mimic the \ac{relu} by preventing eigenvalues from becoming too small, (as the \ac{relu} prevents negative values).
The ReEig function is defined as follows:
\begin{equation}
    \textcolor{black}{C}_k = f_r^{(k)}(\textcolor{black}{C}_{k-1}) = U_{k-1} max(\epsilon I, \Sigma_{k-1}) U_{k-1}^\top
    \label{Eq:ReEig}
\end{equation}
where $k$ is the layer index, $I$ an identity matrix and $\epsilon$ is a threshold of rectification.
An eigenvalue decomposition leads to $U$ and $\Sigma$.
The $max()$ function in this case forces the elements of the diagonal matrix $\Sigma$ to be above the threshold value, $\epsilon$.
This prevents the matrices from getting to close to the singular boundary of the set of positive definite matrices of size $n$, $\mathcal{S}^{pd}_n$.

As in \citep{huangRiemannianNetworkSPD2016}, the BiMap and ReEig layers are applied in pairs, with each successive BiMap reducing the size of the input matrix by half.
The depth of the network is then defined by how many BiMap-ReEig pairs there are, denoted $N_{BiRe}$.

The \textcolor{black}{final Riemannian layer is the} LogEig layer \textcolor{black}{which} transforms the input data into Euclidean space using the \ac{logeuc} \citep{huangRiemannianNetworkSPD2016, arsignyLogEuclideanMetricsFast2006}.
This means a matrix logarithm is applied to each matrix in the input batch.
This is faster than applying the logarithmic map from the \ac{airm}, and there is no need for determining a reference matrix.
The speed difference means it is better suited to the deep learning pipeline, where it will be used for every epoch.

The LogEig function, $f_l$, is defined as follows:
\begin{equation}
    \textcolor{black}{C}_l = f_l^{(k)}(\textcolor{black}{C}_{k-1}) = log(\textcolor{black}{C}_{k-1}) = U_{k-1} log(\Sigma_{k-1}) U_{k-1}^\top
\label{Eq:LogEig}
\end{equation}
where $\textcolor{black}{C} = U\Sigma U^\top$ is an eigenvalue decomposition and $k$ is the layer index.

After this, the output can be passed to a regular fully connected layer for label prediction.

\subsection{End-to-End EEG SPDNet}
\label{Sec: EESPDNet Methods}

\begin{figure}[ht]
    \centering
    \includegraphics[width=\textwidth]{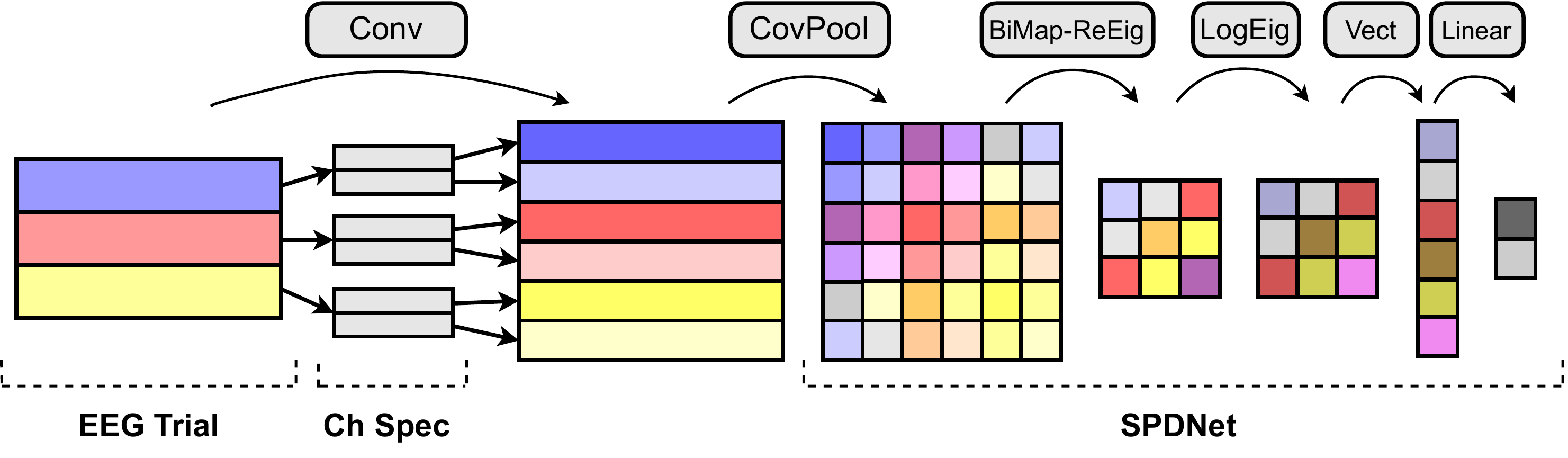}
    \caption[Data pipeline for the \ac{eespdnet}]{
    \textbf{Diagram showing the architecture of the \ac{eespdnet}.} 
    Input trials are convolved in a convolutional layer, which mimics bandpass filtering. 
    An SCM pooling layer turns the convolved time series into sampled covariance matrices, which are \ac{spd}.
    These matrices are then passed into an SPDNet to produce class predictions.
    The \ac{eespdnet} shown here has had certain layers removed to improve clarity, therefore it should be noted that the \ac{eespdnet} used for computations has 3 pairs of BiMap-ReEig layers, and that the BiMap and ReEig layers are separate layers.
    For more detail of the SPDNet components, see the work by \citet{huangRiemannianNetworkSPD2016}.
    }
    \label{Dia:EE-SPDNet Pipeline}
\end{figure}

\textcolor{black}{The \ac{eespdnet} architecture, illustrated in Figure \ref{Dia:EE-SPDNet Pipeline}, incorporates two layers atop the SPDNet for end-to-end processing of \ac{eeg} signals.}
\textcolor{black}{A} convolutional layer similar to that in \citet{schirrmeisterDeepLearningConvolutional2017}, \textcolor{black}{temporally filters the input signals via convolution.}
The \ac{scm} pooling layer then creates \acp{scm} from the convolved \textcolor{black}{(i.e. filtered)} signals and passes them to an SPDNet.
\textcolor{black}{This design allows joint optimisation of the convolutional filterbank with model training.}
\textcolor{black}{
Two types of filter have been used in this study, a regular convolutional filter and a sinc filter.
Both filters filter the incoming signal in the frequency domain via convolution in the time domain, and both use the same length convolutional kernel.
The sinc layer, adapted from an implementation of SincNet \citep{ravanelliInterpretableConvolutionalFilters2019}, generates the convolutional kernel from sinc functions resulting in a band-pass filter frequency response.
Its learned parameters are the low/high cut off frequencies of the band-pass filter.
Conversely the “regular” convolutional filter directly learns the convolutional kernel values.
While the sinc layer is readily interpretable as a band-pass filter, the regular convolutional filter is less restricted, potentially learning sinc filters but also potentially learning more complex filter types.
}

\subsection{Optimised Filterbank EEG SPDNet}
\label{Sec: fbspdnet Methods}


\begin{figure}[t!]
    \centering
    \includegraphics[width=\textwidth]{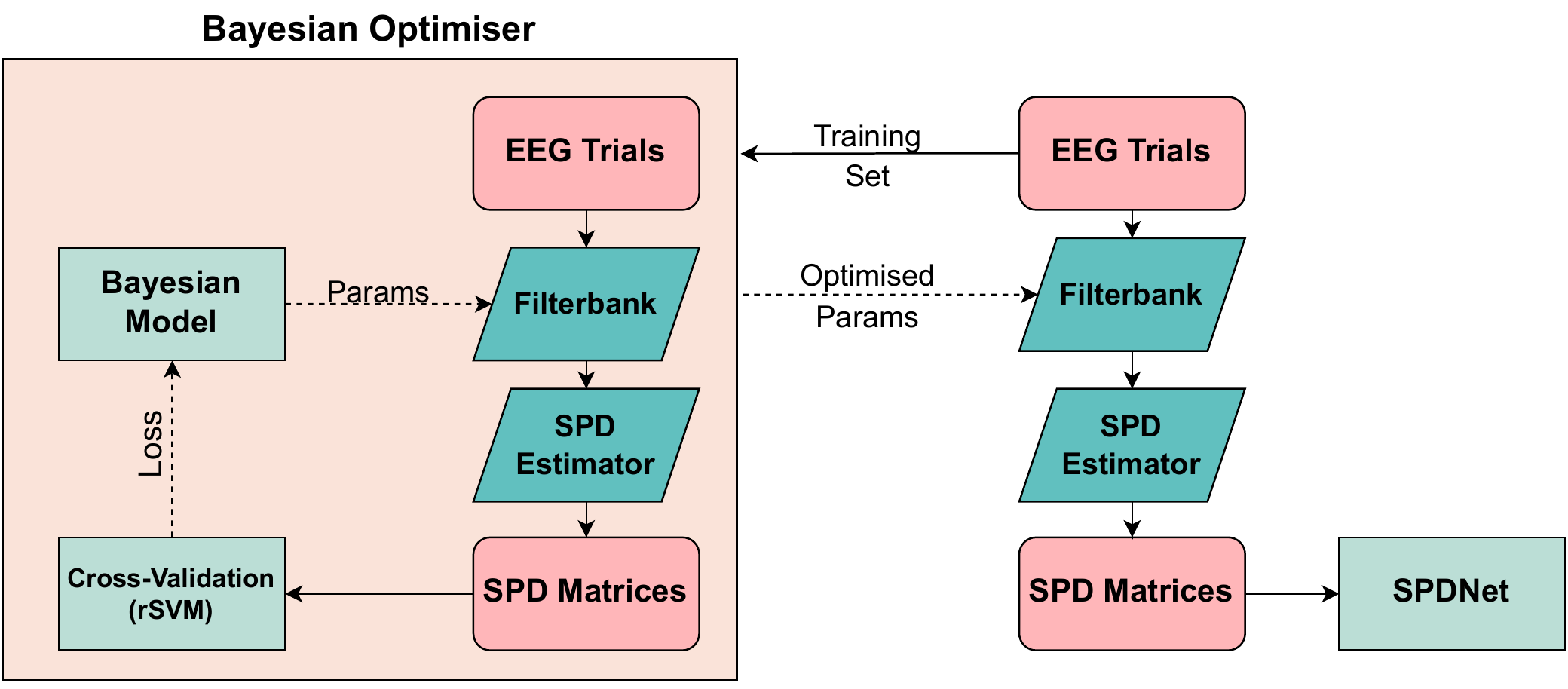}
    \caption[Data pipeline for the BO-SPDNet]{
    \textbf{Diagram showing the architecture of the \ac{fbspdnet}}.
    The training set is passed to a \ac{bo}, where it is used to optimise the filterbank.
    The optimal filterbank is then used for the entire dataset.
    An SPD estimator (in this case the \ac{scm}) is used to create SPD matrices, which are passed to the SPDNet for classification.
    }
    \label{Dia:BO-SPDNetPipeline}
\end{figure}
\textcolor{black}{
FBSPDNet combines a separate \ac{fbopt} loop with covariance matrix generation, followed by a regular SPDNet. 
 It employs a \ac{bo} (although any black box optimisation algorithm could be used) to search the frequency space for the optimal filterbank, as shown in Figure \ref{Dia:BO-SPDNetPipeline}.
}
\textcolor{black}{
The optimiser iteratively adjusts filterbank parameters (bandpass filter cut-offs) to maximise cross-validation accuracy on the computed covariance matrices. 
To reduce computational burden, a lightweight proxy classifier (\ac{svm} or \ac{mdm} with a Riemannian metric) is used instead of training/testing an SPDNet at each iteration. 
Both classifiers were tested with \ac{logeuc} and \ac{airm} Riemannian metrics. 
These Riemannian-based classifiers (\ac{rsvm} and \ac{rmdm}) have been previously used as \acp{rbd} for EEG \citep{barachantClassificationCovarianceMatrices2013, congedoRiemannianGeometryEEGbased2017, congedomarcoRIEMANNIANMINIMUMDISTANCE}.
}
\textcolor{black}{
In order to ensure fair comparisons with Sinc-\ac{eespdnet}, the filterbank for \ac{fbspdnet} was constructed from static sinc-convolution layers.
}

\textcolor{black}{
This optimisation loop, results in a set of easily interpretable filterbank parameters that are optimised to the data.}
The SPD matrices created from the optimised filterbank are passed to an SPDNet.
This SPDNet is identical in architecture to the one in the \ac{eespdnet}.

\subsection{\textcolor{black}{Network} Design Choices}
\label{Sec: Methods/NetworkDesignChoices}

In the following section we elaborate on some design choices specific to the SPDNet, but are applicable to \ac{fbspdnet} and \ac{eespdnet}.

\subsubsection{Optimiser}
\label{Sec: Methods/NetworkDesign/Optimiser}

\textcolor{black}{
We used an implementation \citep{kochurovGeooptRiemannianOptimization2020} of the RiemannianAdam \citep{becigneulRiemannianAdaptiveOptimization2018} optimiser for all training loops involving SPDNet layers.
Aside from learning rate and weight decay, all optimiser parameters were left at default values.
Learning rate and weight decay were found through a coarse grid searches over typical values, see Section \ref{Sec: Methods/Procedure} for more details.
}

\color{black}
\subsubsection{Network Width and Channel Specificity}
\label{Sec: Methods/NetworkDesign/WidthAndChSpec}

Neural network architecture is typically characterised by depth (number of layers) and width (neurons per layer). 
For the proposed SPDNet based models described in this study, \ac{fbspdnet} and \ac{eespdnet}, depth will be defined by the number of BiMap-ReEig pairs in the SPDNet section, $N_{BiRe}$  (see Section \ref{Sec: Rie Overview}), and has a static value of 3 for most of the results seen here.
This section describes how network width is parametrised in this study.

We define the network width, $N_f$, as how many times each input EEG signal gets duplicated during the filtering stage.
So for $N_f > 1$, an individual signal from a single electrode will be filtered multiple times, creating $N_f$ filtered signals per electrode.
Consequently, the number of columns/rows in the covariance matrix size becomes $N_f \times N_e$, which then defines the size of the first BiMap layer and subsequent layers (since the BiMap compression factor is fixed). 
Through the addition of additional weights per layer, increasing $N_f$ potentially enhances each layer's feature-learning capacity.
The models in this study were explored with values of 1 to 8 for $N_f$.

In addition to this, channel specificity was explored at the filtering stage.
In previous works with convolution networks, such as \citep{schirrmeisterDeepLearningConvolutional2017}, and also in typical filterbanks pipelines such as \citep{angFilterBankCommon2012}, each filter is applied to \textit{all} EEG signals.
In this sense the filters are \textit{independent} of the EEG signals.
This study explored \acf{chspec} filtering, where filters are learned for and applied to specific electrodes, potentially extracting task-relevant information at the single-electrode level. 
So, in this study, a \acf{chind} model with $N_f=2$ learns two filters for \textit{all} electrodes (resulting in $N_f$ learned filters) and a \ac{chspec} model with $N_f=2$ learns two filters for \textit{each} electrode (resulting in $N_f \times N_e$ learned filters).
This also means that switching between \ac{chspec} and \ac{chind} filtering for a given $N_f$ does not change the size of the resulting covariance matrix (and therefore does not effect the size of the downstream layers).

\color{black}
\subsection{SPD Matrix Operations}
\label{Sec: SPD matrix ops}
This section will briefly outline some of the SPD Matrix operations that are common to most models/pipelines.

\subsubsection{Vectorisation}
In the "Vect" layer, and when using the \ac{rsvm}, \ac{spd} matrices were vectorised. 
The vectorisation process is as follows:
For $S \in \mathcal{S}^{pd}_n$, the vectorisation operation, $Vect(S)$ stacks the unique elements of $S$ into a $\frac{n}{2}(n+1)$ dimensional vector\footnote{It is also possible to flatten the input matrix entirely, but some early testing suggested keeping redundant information (from the off diagonals) at this stage decreased classification performance.}\textcolor{black}{.}
\textcolor{black}{To preserve equality of norms, a coefficient of $\sqrt{2}$ is applied to the off-diagonals (\cite{barachantClassificationCovarianceMatrices2013}).}
\begin{equation}
    Vect(S) = [S_{1,1}, \textcolor{black}{\sqrt{2}}S_{1,2}, S_{2,2},  \textcolor{black}{\sqrt{2}}S_{1,3},  \textcolor{black}{\sqrt{2}}S_{2,3}, S_{3,3}, ... , S_{n,n}]
    \label{Eq:Vectorisation}
\end{equation}

\subsubsection{Concatenation}
\label{Sec: Concatenation}
When constructing the \ac{spd} matrices from the filterbank parameters (number of \textcolor{black}{filters, $N_f$} > 1) with channel independent filtering, an \ac{spd} matrix was constructed for each \textcolor{black}{filter} (a pair of cut-off frequencies for a bandpass filter).
These \ac{spd} matrices were then concatenated into one larger \ac{spd} matrix, before being passed to the next stage of the pipeline.

The following describes a concatenation procedure for multiple \ac{spd} matrices (of variable size) that results in a matrix that is provably \ac{spd}\footnote{See Supplementary Materials for proof.}.
Let $S_n$ and $S_m$ be \ac{spd} matrices of size $n$ and $m$, respectively.
Let $Conc(S_n, S_m)$ be the mapping: $\mathcal{S}^{pd}_n \times \mathcal{S}^{pd}_m \Rightarrow \mathcal{S}^{pd}_{n + m}$ such that the output is a block diagonal matrix of the form:
\begin{equation}
    Conc(S_n, S_m) = 
    \begin{pmatrix}
    S_n & 0_{n \times m} \\
    0_{m \times n} & S_m \\
    \end{pmatrix}
    \label{Eq:Concatenation}
\end{equation}

The size of the concatenated matrix is equal to the sum of the sizes of the input matrices.
It is trivial to see how this mapping can be consecutively applied to concatenate any number of \ac{spd} matrices.
This allows for the simple concatenation of multiple \ac{spd} matrices, potentially of different size (although this property is not explored here), in a way that retains the \ac{spd} property.

\subsubsection{Interband Covariance}
\label{sec: Interband}
The above described concatenation procedure allows for the combining of \ac{spd} matrices to form a single \ac{spd} matrix.
As previously mentioned, this is used when creating an \ac{spd} matrix for each \textcolor{black}{filter} of a channel independent filterbank.
Therefore, the block diagonal matrix, $C$, may be composed of \ac{spd} matrices $S_1$ \& $S_2$ (\textcolor{black}{both} of size $n$) which are independently computed after filtering trials $T$ through \textcolor{black}{filters $F_1$ \& $F_2$ (i.e.  $S_1$ is the covariance matrix of $T_1$, which is $T$ passed through $F_1$).}
\begin{equation}
    C = Conc(S_1, S_2) = 
    \begin{pmatrix}
    S_1 & 0_{n \times n} \\
    0_{n \times n} & S_2 \\
    \end{pmatrix}
\end{equation}

In $C$, the off diagonal blocks are all 0, however, if $T_1$ \& $T_2$ were first stacked along the electrode axis:
\textcolor{black}{
\begin{equation}
    T' = \begin{pmatrix}
    T_1 \\
    T_2 \\
\end{pmatrix}
\end{equation}
}
one can compute a slightly altered covariance matrix $C'$:
\textcolor{black}{
\begin{equation}
    C' = Cov(T') = 
    \begin{pmatrix}
        S_1 & IC \\
        IC^\top & S_2 \\
    \end{pmatrix}
\end{equation}
}
The off diagonal blocks of $C'$, \textcolor{black}{$IC$,} contain the covariance between electrodes of different \textcolor{black}{frequency} bands, labelled here as the inter frequency band covariance (simply, the interband covariance), and the diagonal blocks contain, as in $C$, the within band covariance.
Setting the off diagonal blocks to $0$ turns $C'$ into $C$.
Both with and without interband covariance models have been explored in this study.

The interband covariance can only be calculated for models with $N_B \geq 2$.
This means that in figures showing singleband results, there is only one channel independent model.

\subsubsection{Regularisation}
\textcolor{black}{
During certain analyses or visualisation, estimated \ac{spd} matrices required regularisation.
When this was needed, matrices were regularised using a ReEig layer.
}
\textcolor{black}{
Additionally, the combination of the BiMap and ReEig layers will force non-\ac{pd} input matrices to be \ac{pd}.
}

\subsection{Analys\textcolor{black}{e}s}
\label{Sec: Methods/Analyses}

In the following section the techniques used for analysis will be detailed.
The main methods of analysis used are those that produce the frequency gain spectra figures and the \ac{lbl} performance assessment.
All analysis was performed using the train/test split from the validation set (i.e. no final evaluation set).

\subsubsection{Chosen Frequency Spectra}
\label{Sec: Methods/Analyses/ChosenFreqs}

\textcolor{black}{
The Sinc layer used in \ac{fbspdnet} and Sinc-\ac{eespdnet} offers easily interpretable filterbank parameters as sets of high- and low-pass filter cutoff frequencies.
When displaying this information, such as in Figure \ref{Fig:FBSPD ChosenFreqs}, we show the frequency coverage on the y-axis.
This is a percentage value that indicates how many of the bandpass filters of that group of models contain that particular frequency, so 75\% at 30Hz indicates that 75\% of bandpass filters covered 30Hz, for that model grouping.
}

\subsubsection{Frequency Gain Spectra}
\label{Sec: FreqGainSpec Method}
\textcolor{black}{Visualising the learned filterbank of} \ac{eespdnet} is not so simple, as it uses convolutional filters, which do not directly correspond to a particular region in the frequency space.
\textcolor{black}{To assess and visualise the frequencies filtered by the convolutional layer, we calculated the frequency gain spectra by comparing pre- and post-filterbank/convolution frequency spectra.}
\textcolor{black}{The frequency gain, shown in dB, represents amplification (positive values) or attenuation (negative values) of frequencies relative to the original input (Figure \ref{Fig:EE-SPDNet Freq Gain Spectra}). }

\textcolor{black}{The average frequency gain spectra seen in the figures here averages across participants, seeds, and filters, and t}he precise order of calculations and operations performed for the gain spectra is detailed in the Supplementary Materials (Section \ref{Sec: Appdx FreqGainSpec}).
\textcolor{black}{In some cases, certain frequency regions were attenuated so much that their frequency gain became negative infinity.
Negative infinity values were usually interpolated, but a handful of spectra with over 50\% negative infinities were discarded before plotting.}

\color{black}
\subsubsection{Peak Detection for Multiband Assessment}
\label{Sec: Methods/Analyses/Peak Detection}

The learned filters of \ac{eespdnet} were analysed to count how many would be classed as “multiband” filters.
As is explained in more detail in Sections \ref{Res: F5: Multiband} \& \ref{Disc: Filter Type} a multiband filter is a bandpass filter that lets two or more distinct frequency regions pass.
Examples of multiband filters can be seen in the Electrode-Frequency relevance plots in Figure \ref{Fig: CovgradP12}.

To analyse the occurrence of multiband filters in EEGSPDNet's learned filterbanks, we employed a peak detection algorithm on the frequency gain spectra.
After discarding/interpolating troublesome spectra (as previously explained), the signals were smoothed with a moving average and zeroed around the median.
We then used the an out-of-the-box peak detection function (see Supplementary Materials for details) with heuristically set height and width parameters to detect peaks while avoiding noise.

\subsubsection{Electrode-Frequency Relevance}
\label{Sec: Methods/Analyses/Electrode Relevance}

Electrode-frequency relevance, shown in Figures \ref{Fig: CovgradP5} \& \ref{Fig: CovgradP12}, combines frequency gain spectra and electrode relevance. 

The relevance values for these plots are aggregated gradients calculated from the feature map after the covariance pooling layer, and before the first BiMap layer (i.e. the computed covariance matrix). 
Gradients of each prediction with respect to the associated covariance matrix in the training set were calculated, summed across rows, and multiplied by the associated normalised frequency gain spectra. 
This process, while computationally intensive, provides valuable insights into the features discussed in this study.
\color{black}

\subsubsection{Layer-by-Layer Performance}
\label{Sec: Methods/Analyses/LBL}

In order to provide some insight into architectural performance, the network was analysed in a \acf{lbl} manner.
This involved passing the data through the trained network, and classifying at every suitable stage in the network.
The same preprocessing and data splitting was used here, such that the data exposed to the classifier is the same as the data used by the network.
This allows for an understanding of the effect of each layer on classification performance as well as an evaluation of network architecture which may lead to possible areas of improvement.
Classification was performed with a Euclidean (SVM) and a Riemannian (\ac{rsvm}) classifier, to give insight into the data in each of these spaces.
Furthermore this may lead to some insights into the internal representation of the data, which is explored in some visualisations.
\color{black}
\subsubsection{BiMap Gain}
\label{Sec: Methods/Analyses/Bimap Gain}

In order to visualise the usage of spatial information by the trained networks, we aggregated the weights of the first BiMap layer.
This is a linear layer which takes the computed \ac{scm} as input.
Therefore, it is possible to calculate the contribution of each term in the \ac{scm} to the output.
We label the contribution of the input to the output the gain.

For a \ac{scm} $C$ of size $N_C$, output matrix $M$ of size $N_M$, and BiMap weights $W$ we can write equation \ref{Eq:BiMap} as:
\begin{equation}
    M = W^\top C W
\end{equation}
Therefore, the $i^{th}$ and $j^{th}$ component of the output, $M_{ij}$ is:

\begin{equation}
    M_{ij} = \sum^{N_C}_{p=1} \sum^{N_C}_{q=1} W^\top_{ip} C_{pq} W_{qj}
\end{equation}

The gain applied to $C_{pq}$ for an output $M_{ij}$ is then given by:
\begin{equation}
    G_{pq}^{ij} = W^\top_{ip} W_{qj}
\end{equation}

Therefore, the gain applied to $C_{pq}$ across the whole of $M$ is:
\begin{equation}
    G_{pq} = \sum^{N_C}_{i=1} \sum^{N_C}_{j=1} W^\top_{ip} W_{qj}
\end{equation}
which is the summed outer product of the $p^{th}$ and $q^{th}$ rows in $W$:

\begin{equation}
    G_{pq} = \sum^{N_C}_{i=1} \sum^{N_C}_{j=1} (W_p \otimes W_q)_{ij}
\end{equation}

The symmetric matrix $G$ then shows the gain applied to each entry in C.
To visualise the spatial gain, we summed G along the rows.
This means that the "single electrode" values seen in the plots are the summation of the gains applied to that variance of that electrode plus the gain applied to it's covariance with every other electrode.

\subsection{Comparison Models}
\label{Sec: Methods/ComparisonModels}

We have implemented 4 other deep networks for EEG to serve as comparisons to our models in final evaluation.
All are deep convolutional networks, and as such also learn convolutional filters (i.e. they learn the filterbanks) during training. 
Deep4Net \citep{schirrmeisterDeepLearningConvolutional2017}, ShallowFBCSPNet \citep{schirrmeisterDeepLearningConvolutional2017} and EEGNetv4 \citep{lawhernEEGNetCompactConvolutional2018} are all standard convolutional networks with differing architectures. 
TSMNet \citep{koblerSPDDomainspecificBatch2022} differs from the others in that it is a SPDNet based \ac{drn}, originally designed for domain adaptation tasks.
The key differences between TSMNet and our proposed models are that  $N_{BiRe}=1$ for TSMNet and it uses a Riemannian batch normalisation layer, as well as a similar convolutional filtering stage to Deep4Net and ShallowFBCSPNet (temporal followed by spatial filtering).

For Deep4Net, we have not implemented cropped decoding, which will likely decrease it's classification performance.
Regarding TSMNet, we have used the SPD momentum batch normalisation layer\citep{koblerSPDDomainspecificBatch2022}, as it was most appropriate to our setting.
For other technical aspects of these models we would direct the reader to their associated papers.
\color{black}
\subsection{Datasets}
\label{Sec: Methods/Datasets}

We used \textcolor{black}{5} public \textcolor{black}{motor EEG} datasets for this study: \textcolor{black}{Schirrmeister2017} \citep{schirrmeisterDeepLearningConvolutional2017}\textcolor{black}{, BNCI2014\_001 \citep{tangermannReviewBCICompetition2012}, BNCI2014\_004 \citep{leebBrainComputerCommunication2007}, Lee2019\_MI \citep{leeEEGDatasetOpenBMI2019} and Shin2017A \citep{shinOpenAccessDataset2017}.
All of which were downloaded using MOABB \citep{jayaramMOABBTrustworthyAlgorithm2018}.
Table \ref{Tab: Datasets} in the Supplementary Materials lists various data specific details.
In general, the datasets underwent minimal preprocessing: scaling, clipping, resampling and a single bandpass filter.}
\color{black}
For final evaluation we used the default train/test splits of all datasets.
During hyperparamter exploration (i.e. the validation phase) we used a sequential 80:20 split of the training set from Schirrmeister2017.
The holdout set/final evaluation set of this dataset was kept "unseen".

As will be further detailed in the following section, since much of the analysis explores how results vary with different hyperparameter combinations, all of the analysis was conducted on this validation split.

\subsection{Procedure}
\label{Sec: Methods/Procedure}

This section outlines the experimental procedure, including training loops, data splits, and hyperparameter searches.
The results are categorized into two phases: a hyperparameter search phase and a final evaluation phase.

The hyperparameter search phase utilized a validation split of the Schirrmeister2017 dataset to determine learning rate, weight decay, and network design parameters. 
Analysis was applied to models from this phase to examine how the data changed with these parameters. 
The best model of each proposed design was selected for the final evaluation phase.
In this phase, the models were trained (separately) on the 5 datasets.
All datasets were downloaded using MOABB \citep{jayaramMOABBTrustworthyAlgorithm2018}, and had minimal preprocessing applied.

For either phase, every training/testing loop was applied in a \textit{subject-independent} manner and with 3 random seeds.
This means that for any given dataset, the network would be trained/tested on a single participant 3 times, each time with a different random initialisation. 
Scores/analyses were then averaged across participants \& seeds to give a single accuracy for that model on that dataset.
This approach is also used for the filterbank search done in the \ac{fbspdnet} pipeline, resulting a 3 filterbanks per participant (one per seed).

FBSPDNet filterbanks were searched using Bayesian optimization with four objective functions based on two metrics (\ac{airm} and \ac{logeuc}) and two classifiers (rMDM and rSVM). 
The search was limited to 1000 trials or 12 hours of walltime.
The resulting filterbanks were then used to filter the data before covariance matrix computation and SPDNet training.
Optimization was performed on the training set only, making \ac{cv} accuracies from \ac{fbopt} classification with \ac{rmdm} or \ac{rsvm} not directly comparable to SPDNet scores.

\begin{figure}
    \centering
    \includegraphics[width=\textwidth]{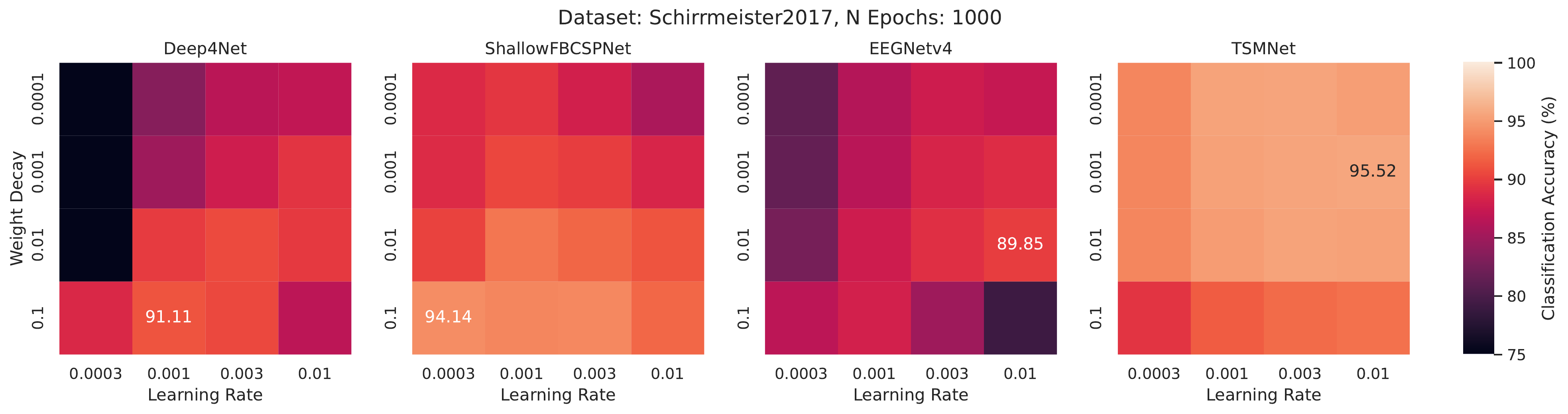}
    \caption{
    \textcolor{black}{
\textbf{Learning Rate \& Weight Decay Parameter Sweep for Comparison Models.}
    Each heatmap subplot shows the coarse parameter sweep performed for selecting weight decay and learning rate for each model.
    The model is displayed as the heatmap title, see Section \ref{Sec: Methods/ComparisonModels} for details on the comparison models.
    Weight decay values are shown on the y-axis, learning rate on the x-axis, with hue indicating the models classification performance on the test set (of the validation set, see Section \ref{Sec: Methods/Procedure}.
    The cell with the highest classification accuracy has been annotated.
    }}
    \label{Fig: ParamSweep ComparisonModels}
\end{figure}

\begin{figure}
    \centering
    \includegraphics[width=\textwidth]{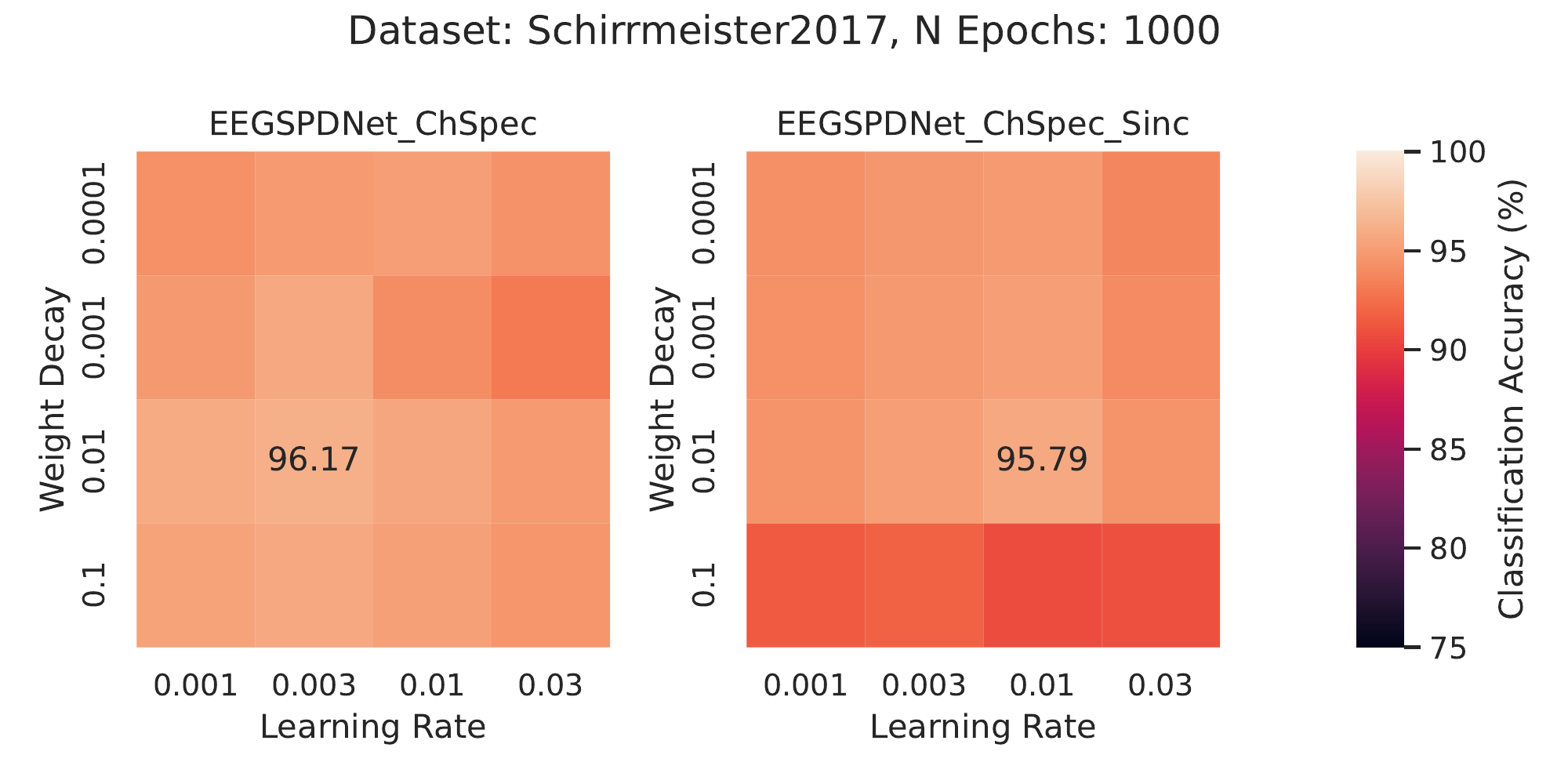}
    \caption{
    \textcolor{black}{
\textbf{Learning Rate \& Weight Decay Parameter Sweep for EEGSPDNet}
    See Figure \ref{Fig: ParamSweep ComparisonModels} for figure structure details.
    Data shown is for \ac{eespdnet} \ac{chspec} $N_f=8$, for conv (left) and sinc (right) filtering.
    }}
    \label{Fig: ParamSweep EEGSPDNet}
\end{figure}

\begin{figure}
    \centering
    \includegraphics[width=\textwidth]{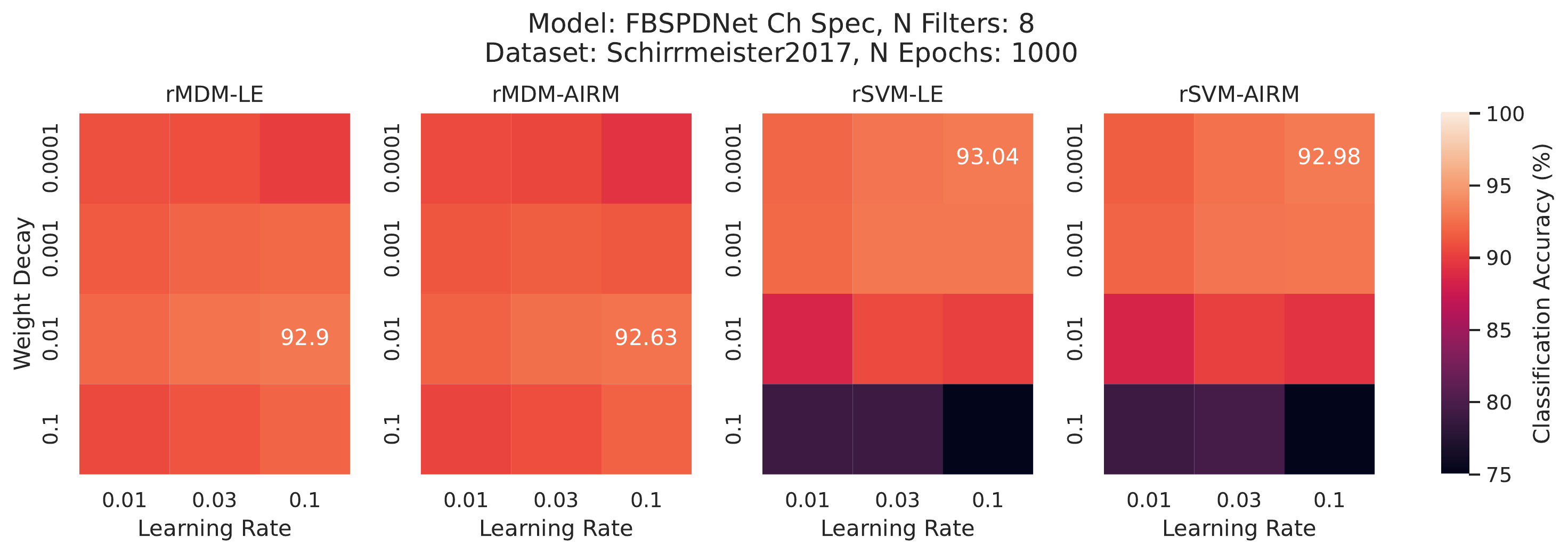}
    \caption{
    \textcolor{black}{
\textbf{Learning Rate \& Weight Decay Parameter Sweep for FBSPDNet}
    See Figure \ref{Fig: ParamSweep ComparisonModels} for figure structure details.
    Data shown is for \ac{fbspdnet} \ac{chspec}, $N_f=8$, for all the different proxy classifier configurations during filterbank optimisation.
    There are two classfiers, rMDM and rSVM, and two Riemannian metrics, LE and AIRM.
    }}
    \label{Fig: ParamSweep FBSPDNet}
\end{figure}

Values for the weight decay and learning rate were found via coarse grid search of typical values with a subset of models.
The first gridsearch was used to choose a static value for network depth, parametrised by $N_{BiRe}$, and can be seen in Figure \ref{Fig: ParamSweep Depth}.
The search used an 8-filter \ac{chspec} \ac{eespdnet}, and a value of $N_{BiRe}=3$ was brought forwards for all other variants of \ac{fbspdnet} and \ac{eespdnet}.
A Sinc variant of the same model was used in another search to set the values of learning rate and weight decay for the \ac{eespdnet} models (Figure \ref{Fig: ParamSweep EEGSPDNet}).
Next, weight decay and learning rate values for the 4 \ac{fbspdnet} variants were also searched, using a Channel specific 8 Filter model (see figure \ref{Fig: ParamSweep FBSPDNet}), these values were kept for any other variants of the \ac{fbspdnet}.
Finally, the weight decay and learning rates for the 4 comparison models were also searched (see Figure \ref{Fig: ParamSweep ComparisonModels}).
These are the learning rate and weight decay values used for the remainder of computations.

The proposed models were then trained/tested for a range of network widths, $N_f \in \{1, 2, 3, 4, 5, 6, 7, 8\}$ (much of the analysis was performed on the data acquired during this stage).
It is with these results that the best \ac{eespdnet} and best \ac{fbspdnet} were chosen and brought forwards to final evaluation.
They were then trained/tested on the final evaluation datasets using the hyperparameters found previously, and compared against the other out-of-the-box models.

For both  phases, classification accuracy refers to unseen test set performance, with the exceptions of the tSNE plots (Figures \ref{Fig: TSNE Ind} \& \ref{Fig: TSNE Spec}) and the electrode-frequency relevance plots (Figures \ref{Fig: CovgradP5} \& \ref{Fig: CovgradP12}) where both training set and test set accuracies are given.

\subsection{Statistics}
\label{Sec: Methods/Statistics}

In some results/figures p-values are given to assess the significance of the model comparisons. 
For these tests, the Wilcoxon signed-rank test (see Section \ref{Appdx: Software} for package details) was used.

For the heatmap displayed in Figure \ref{Fig: EvalSet_EstPlot_SigMatrix}, accuracies were averaged across seeds.
It is from these scores that the mean differences and p-values were calculated.

For Tables \ref{Tab: ScatterCompTable Ind vs Spec}, \ref{Tab: ScatterCompTable FBSPD vs EEGSPD} and \ref{Tab: ScatterCompTable Interband}, which respectively pair with the left, centre and right subplots of Figure \ref{Fig: EEvsBO_SpecVsInd_RMINT}, values were averaged across seeds and participants to give a single accuracy per value of $N_f$.
It is from these scores that means, standard deviations and p-values were calculated.

In other results, we display model comparisons using estimation plots, using the python package `dabest` \citep{hoMovingValuesData2019}.
\color{black}

\section{Results}
\label{Results}
\color{black}
In the following section we detail our experimental results in 8 findings.
The first 3 of which are focussed on the classification performance between models.
The remaining 5 findings concern patterns in network behaviour that we observed through our analyses.
Lastly, we present some visualisations of some specific data, which demonstrates interesting trends, but does not necessarily possess the same robustness as the previous findings.

\subsection{Finding 1: Deep Riemannian Networks with Learnable Filterbanks Can Outperform ConvNets}
\label{Res: F1: Main Accs}

\begin{figure}[ht]
    \centering
    \includegraphics[width=\textwidth]{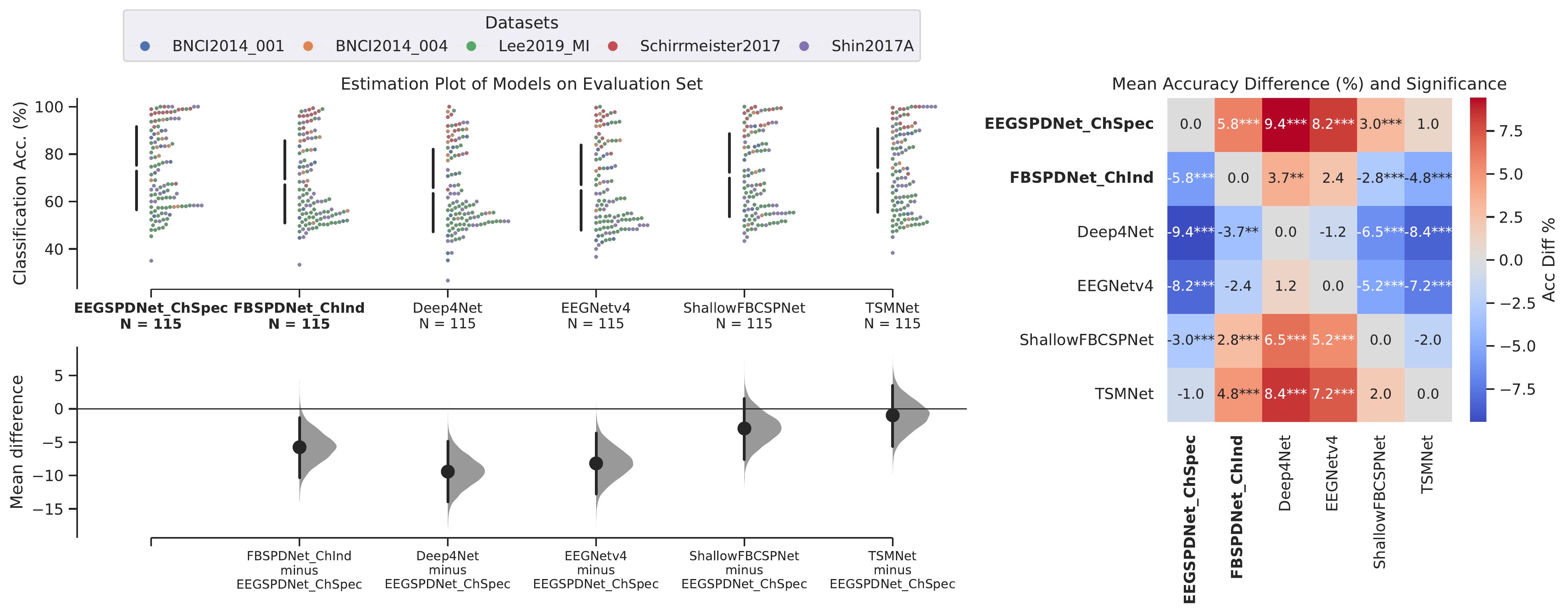}
    \caption{\color{black}
\textbf{\textcolor{black}{Final Evaluation Set} Results.}
These figures show the results of our models (EEGSPDNet and FBSPDNet) and the comparison models on the evaluation datasets.
Details regarding the comparison models we used can be found in Section \ref{Sec: Methods/ComparisonModels}, details regarding the datasets can be found in Section \ref{Sec: Methods/Datasets}.
As discussed previously, the variants of our proposed models are an 8-filter, channel specific \ac{eespdnet} (EEGSPDNet \ac{chspec}) and a 6-filter channel independent \ac{fbspdnet} (FBSPDNet ChInd).
One the left is an estimation plot, which is itself made up of two subplots.
The upper subplot is a swarm plot, with the models on x-axis, and the test-set classification accuracy on the y-axis.
Each point represents a single participant training-testing loop from a single dataset (hue denotes which dataset) for a given model.
To the left of a particular models swarm is the is a gapped line showing the swarm mean +/- standard deviation.
There are 115 points in each swarm, which is the number of participants across all datasets.
The lower subplot shows the bootstrapped ($n=10000$) mean difference between the left-most model (EEGSPDNet \ac{chspec}) and every other model.
The shaded area shows the distribution of the bootstrapped differences, with the dot and line respectively showing the mean an 95\% confidence intervals.
On the right is a heatmap displaying mean accuracy differences (not bootstrapped, as in the estimation plots) and asterisks for significance thresholds (one, two or three asterisks imply significance less than 0.05, 0.01 \& 0.001, respectively).
A cell in the heatmap represents the row minus column accuracy difference.
    }
    \label{Fig: EvalSet_EstPlot_SigMatrix}
\end{figure}

Figure \ref{Fig: EvalSet_EstPlot_SigMatrix} and Table \ref{Tab: EvalAccs} illustrate the evaluation set performance of the best performing \ac{eespdnet} and \ac{fbspdnet} models across 5 datasets (see Section \ref{Sec: Methods/Datasets}), and compared with 4 other models (see Section \ref{Sec: Methods/ComparisonModels}).
EEGSPDNet demonstrated superior performance, achieving statistically significant improvements in classification accuracy against the \ac{fbspdnet} model, Deep4Net, ShallowFBCSPNet, and EEGNet, with a small non-significant increase against TSMNet.
In contrast, \ac{fbspdnet} showed more modest improvements, outperforming only Deep4Net and EEGNet, with only the former at some statistically significant level.
Overall, the \acp{drn} (EEGSPDNet, \ac{fbspdnet} and TSMNet) generally outperformed the non-Riemannian models. 
Additionally, ShallowFBCSPNet was the best performing non-Riemannian model, achieving a statistically significant accuracy increase against \ac{fbspdnet}, Deep4 and EEGNet.
Lastly, TSMNet's performance was marginally below EEGSPDNet's, with slightly less statistical significance against the other models.

\begin{table}
\begin{threeparttable}
\caption{
\textbf{\textcolor{black}{Final Evaluation Set Accuracies}.}
\textcolor{black}{Left-most column shows the} \textcolor{black}{dataset name.
Each subsequent column shows the final evaluation set classification accuracy achieved by that model.
Values have been rounded to 3 significant figures.
For spacing reasons the following model names have been abbreviated, in comparisons to Figure \ref{Fig: EvalSet_EstPlot_SigMatrix}, which is showing the same underlying data: \ac{eespdnet} \ac{chspec} > \ac{eespdnet}, \ac{fbspdnet} ChInd > \ac{fbspdnet} and ShallowFBCSPNet > ShFBCSPNet.
Precise details on the models and datasets and be found in Section \ref{Sec: EESPDNet Methods} and Section \ref{Sec: fbspdnet Methods} for \ac{eespdnet} and \ac{fbspdnet}, Section \ref{Sec: Methods/ComparisonModels} for the comparison models and Section \ref{Sec: Methods/Datasets} for datasets.}
\label{Tab: EvalAccs}}
{\color{black}\begin{tabular}{@{}lllllll}
\toprule
Dataset & \textbf{EEGSPDNet} & \textbf{FBSPDNet} & Deep4Net & EEGNetv4 & ShFBCSPNet & TSMNet  \\
\midrule
     BNCI2014001 & 74.4 & 66.5 & 63.0 & 64.3 & 72.5 & 74.9 \\
     BNCI2014004 & 77.8 & 74.8 & 79.6 & 81.4 & 78.8 & 75.2 \\
     Lee2019 MI & 67.9 & 62.3 & 59.5 & 62.2 & 66.7 & 65.9 \\
     Schirrmeister2017 & 95.2 & 92.8 & 88.1 & 91.2 & 94.0 & 94.5 \\
     Shin2017A & 74.0 & 66.3 & 58.9 & 56.1 & 65.5 & 74.9 \\
\bottomrule
\end{tabular}}
\end{threeparttable}
\end{table}

\subsection{Finding 2: Highest Performance Achieved by a Wide, Convolutional, Channel Specific, DRN}
\label{Res: F2: Wide Conv ChSpec}

\begin{figure}[ht]
    \centering
    \includegraphics[width=\textwidth]{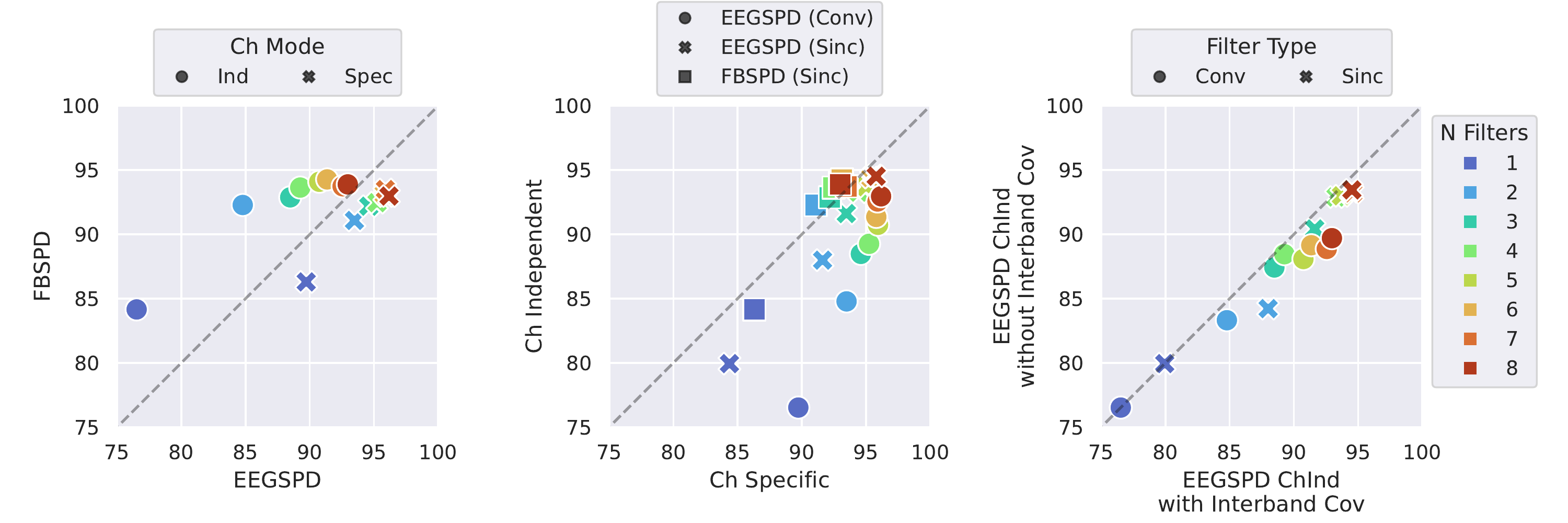}
    \caption{\color{black}
    \textcolor{black}{
    \textbf{Validation Set Classification Accuracy Comparison For Different Conditions}
    From left to right the subplots show:} \ac{fbspdnet} \textcolor{black}{against} \ac{eespdnet} with marker denoting \ac{chind} or \ac{chspec} filtering, \ac{chind} against \ac{chspec} filtering with marker denoting \ac{eespdnet} or \ac{fbspdnet} and removed interband covariance against with interband covariance with markers denoting regular conv filtering or sinc-conv filtering.
    \textcolor{black}{Data has been averaged across seeds and participants, and separated by number of} filters \textcolor{black}{(via colour).}
    Means, standard deviations and p-values for the three subplots can be found in Tables \ref{Tab: ScatterCompTable FBSPD vs EEGSPD}, \ref{Tab: ScatterCompTable Ind vs Spec} and \ref{Tab: ScatterCompTable Interband}.
    The similarity between many p-values arises from the relatively small sample size (7 or 8 values) and that the ranked sums are often identical (i.e. all points are in favour of a particular condition).
    }
    \label{Fig: EEvsBO_SpecVsInd_RMINT}
\end{figure}

The best performing model overall, was an 8 filter, channel specific \ac{eespdnet} using a standard convolutional filter.
It achieved the highest scores on the validation set (see Figure \ref{Fig: EEvsBO_SpecVsInd_RMINT} (left)), and was therefore brought forward to the evaluation set phase on 5 datasets, where it also achieved the highest accuracies (see Figure \ref{Fig: EvalSet_EstPlot_SigMatrix}).
The highest performing \ac{fbspdnet} model on the validation set (see Figure \ref{Fig: EEvsBO_SpecVsInd_RMINT} (left)) was a 6 filter channel independent model, although it failed to achieve notable performance on the evaluation datasets.
While these two models had differing preferences for channel specificity, both showed that generally wider networks (i.e. more filters) perform better (see Figure \ref{Fig: EEvsBO_SpecVsInd_RMINT} (left \& centre)).
However, this effect appears to plateau for \ac{fbspdnet} at $N_f=6$ (see Figure \ref{Fig: EEvsBO_SpecVsInd_RMINT} (left)).

\begin{table}
\begin{threeparttable}
\caption{
\textcolor{black}{
\textbf{FBSPDNet vs \ac{eespdnet} Averaged Across $N_f$}
This table shows mean +/- std values for the data displayed in Figure \ref{Fig: EEvsBO_SpecVsInd_RMINT} (left).
}
\label{Tab: ScatterCompTable FBSPD vs EEGSPD}}
{\color{black}\begin{tabular}{@{}llll}
\toprule
Filter Specificity & \ac{eespdnet} (Conv) & \ac{fbspdnet} & p-value \\
\midrule
     Channel Independent & 89.8 +/- 5.2 & 92.4 +/- 3.4 & 0.0078125 \\
     Channel Specific & 93.9 +/- 3.0 & 91.8 +/- 2.6 & 0.0078125 \\
\bottomrule
\end{tabular}}
\end{threeparttable}
\end{table}

As previously stated, the best performing model was the 8 filter Conv-\ac{eespdnet} with channel specific filtering.
In fact, the Conv-\ac{eespdnet} outperformed the Sinc-\ac{eespdnet} and the 4 \ac{fbspdnet} variants across all $N_f$ in the \ac{chspec} setting (Figure \ref{Fig: ChSpec EstPlot}).
However, this trend was reversed in the \ac{chind} setting (Figure \ref{Fig: ChInd EstPlot}, where the Conv-\ac{eespdnet} was outperformed by the Sinc-\ac{eespdnet}, and both were beaten by some variants of the \ac{fbspdnet} (the ones that used the rSVM as their proxy classifier for the \ac{fbopt}).
These results suggest that both \ac{chspec} filtering and convolutional filters were crucial for achieving the highest decoding accuracies.

\begin{table}
\begin{threeparttable}
\caption{
\textcolor{black}{
\textbf{\ac{chind} vs \ac{chspec} Averaged Across $N_f$}
This table shows mean +/- std values for the data displayed in Figure \ref{Fig: EEvsBO_SpecVsInd_RMINT} (center).
}
\label{Tab: ScatterCompTable Ind vs Spec}}
{\color{black}\begin{tabular}{@{}llll}
\toprule
Model & Channel Independent & Channel Specific & p-value \\
\midrule
     Conv-\ac{eespdnet} & 88.3 +/- 5.4 & 94.6 +/- 2.2 & 0.0078125 \\
     Sinc-\ac{eespdnet} & 91.3 +/- 5.1 & 93.2 +/- 3.8 & 0.0078125 \\
     \ac{fbspdnet} & 92.4 +/- 3.4 & 91.8 +/- 2.4 & 0.1953125 \\
\bottomrule
\end{tabular}}
\end{threeparttable}
\end{table}

\subsection{\textcolor{black}{Finding 3: Including Covariance between Frequency Bands can Boost Accuracy}}
\label{Res: F3: Interband}

As can be seen in Figure \ref{Fig: EEvsBO_SpecVsInd_RMINT} (right) the removal of the interband covariance negatively affected performance for both Conv- and Sinc- \ac{eespdnet}.
This effect appeared strongest in the Conv-\ac{eespdnet}.

\begin{table}
\begin{threeparttable}
\caption{
\textcolor{black}{
\textbf{With vs Without Interband Covariance Averaged Across $N_f$}
This table shows mean +/- std values for the data displayed in Figure \ref{Fig: EEvsBO_SpecVsInd_RMINT} (right).
}
\label{Tab: ScatterCompTable Interband}}
{\color{black}\begin{tabular}{@{}llll}
\toprule
EEGSPDNet Filter Type & With Interband & Without Interband (RmInt) & p-value \\
\midrule
     Convolution & 88.3 +/- 5.4 & 86.4 +/- 4.5 & 0.015625 \\
     Sinc & 91.3 +/- 5.1 & 90.1 +/- 5.1 & 0.015625 \\
\bottomrule
\end{tabular}}
\end{threeparttable}
\end{table}

\subsection{\textcolor{black}{Finding 4: Learned Frequency Profile is Physiologically Plausible}}
\label{Res: F4: Frequencies}
\begin{figure}[ht!]
    \centering
    \includegraphics[width=\textwidth]{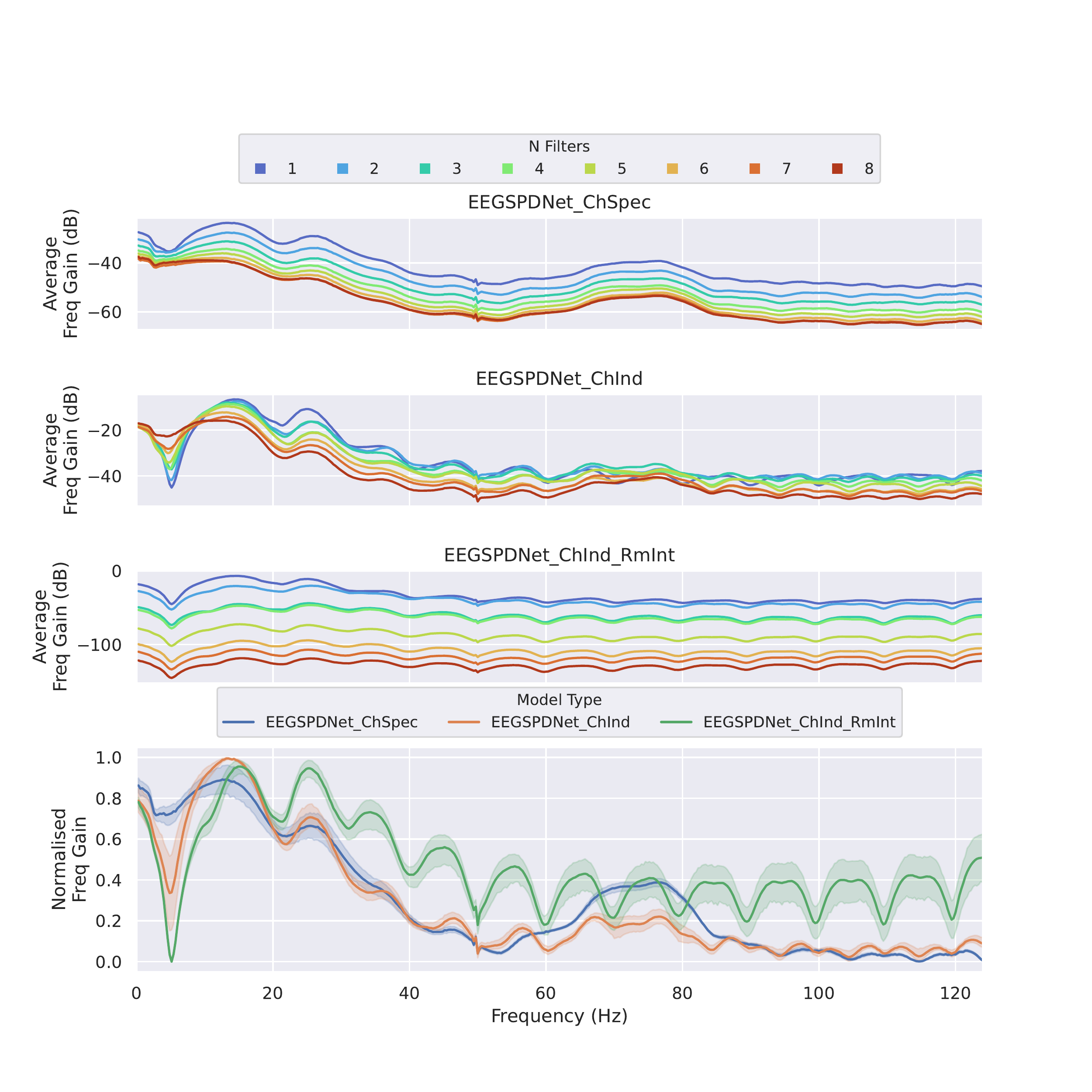}
    \caption{\textcolor{black}{
    \textbf{Average Frequency Gain Caused by Learned Convolutional Filters of \ac{eespdnet}.}
    Each subplot shows the average frequency gain (in decibels) across the spectrum.
    The first three subplots (from top to bottom) show the different model sub-types, with colour denoting the number of filters.
    The bottom subplot shows the average (with standard deviation in shading) across all filters for each model sub-type, frequency gains were normalised between 0 - 1 to allow for better inter-model comparisons.
    All data used for the generation of this figure was collected during the validation phase.
    }}
    \label{Fig:EE-SPDNet Freq Gain Spectra}
\end{figure}

\begin{figure}[ht]
    \centering
    \includegraphics[width=\textwidth]{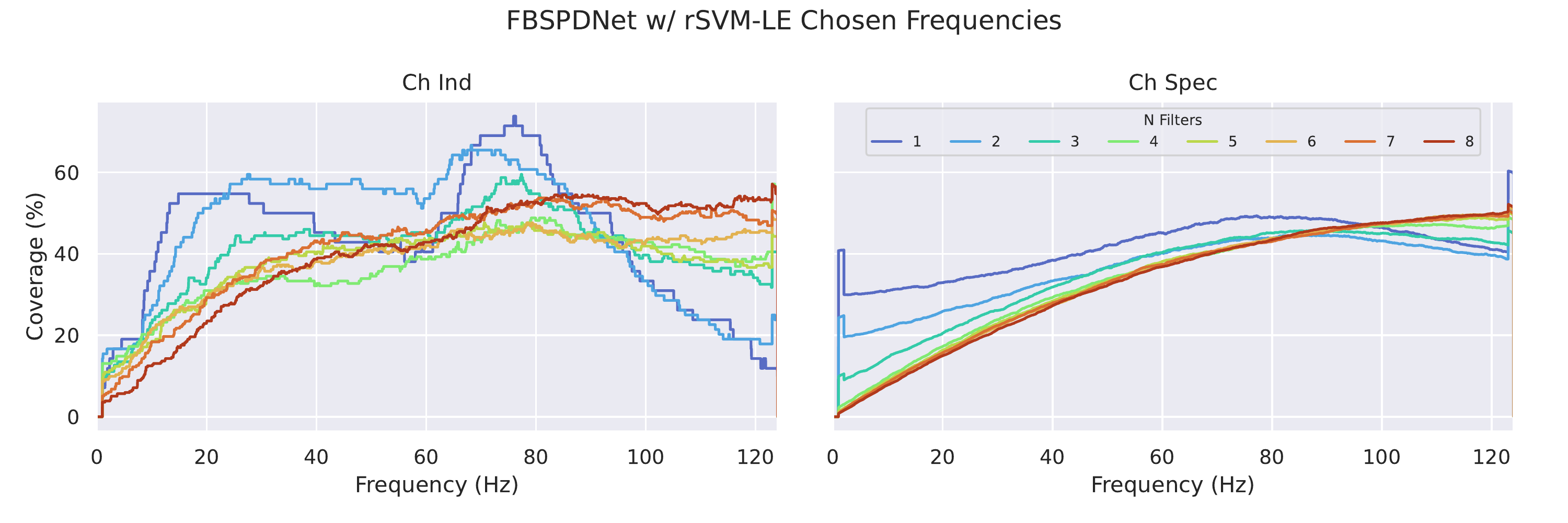}
    \caption{
    \textbf{Frequency Band Selection Distributions for \ac{fbspdnet}.}
    Each subplot shows the selected frequency distribution across the spectrum\textcolor{black}{, left for \ac{chind} right for \ac{chspec}}.
    The colour of the line denotes the number of \textcolor{black}{filters} for that model.
    A value of 100\% means that frequency was included in every bandpass filter (including every participant and if channel specific, and every electrode)\textcolor{black}{ see Section \ref{Sec: Methods/Analyses/ChosenFreqs} for precise details.}
    All data used for the generation of this figure was collected during the validation phase.
    }
    \label{Fig:FBSPD ChosenFreqs}
\end{figure}

The frequencies selected by the optimised filterbanks were also analysed.
Frequency gains resulting from the convolutional filter training (Figure \ref{Fig:EE-SPDNet Freq Gain Spectra}) and the frequency regions selected by the optimizer in the FB-Opt stage of the \ac{fbspdnet} (Figures \ref{Fig:FBSPD ChosenFreqs} and Section \ref{Sec: Appdx: Chosen Freqs}) exhibited prominent peaks in three distinct ranges: 8-20 Hz, 20-35 Hz, and 65-90 Hz.
These ranges correspond to the alpha, beta, and high-gamma bands, respectively.
Notably, classifiers incorporating features from the high-gamma band demonstrated superior performance overall.
\textcolor{black}{These areas are consistent with the areas activated by motor movement \citep{ballMovementRelatedActivity2008, pfurtschellerMotorImageryDirect2001}.}
Additionally, manual inspection of particular electrodes in the electrode-frequency relevance plots (Figures \ref{Fig: CovgradP5} \& \ref{Fig: CovgradP12}), suggests that the networks also select physiologically plausible spatial regions.

\begin{figure}[ht!]
    \centering
    \includegraphics[width=\textwidth]{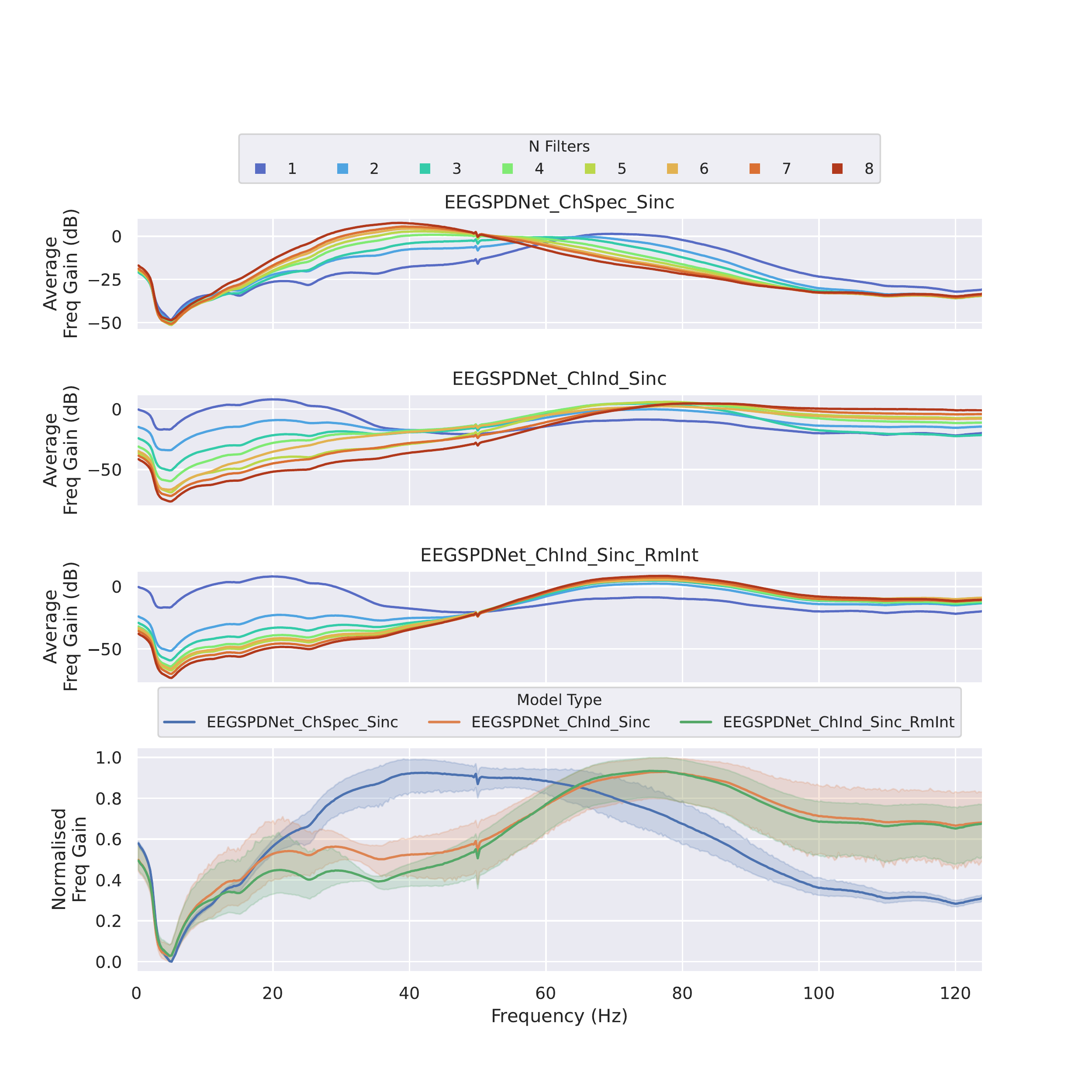}
    \caption{\textcolor{black}{
    \textbf{Average Frequency Gain Caused by Learned Convolutional Filters of \ac{eespdnet}.}
    Each subplot shows the average frequency gain (in decibels) across the spectrum.
    The first three subplots (from top to bottom) show the different model sub-types, with colour denoting the number of filters.
    The bottom subplot shows the average (with standard deviation in shading) across all filters for each model sub-type, frequency gains were normalised between 0 - 1 to allow for better inter-model comparisons.
    All data used for the generation of this figure was collected during the validation phase.
    }}
    \label{Fig:EE-SPDNet Sinc Freq Gain Spectra}
\end{figure}
\color{black}

\color{black}
\subsection{Finding 5: Conv-EEGSPDNet Models may Learn Task-Relevant Multiband Filters}
\label{Res: F5: Multiband}

\begin{figure}[ht]
    \centering
    \includegraphics[width=\textwidth]{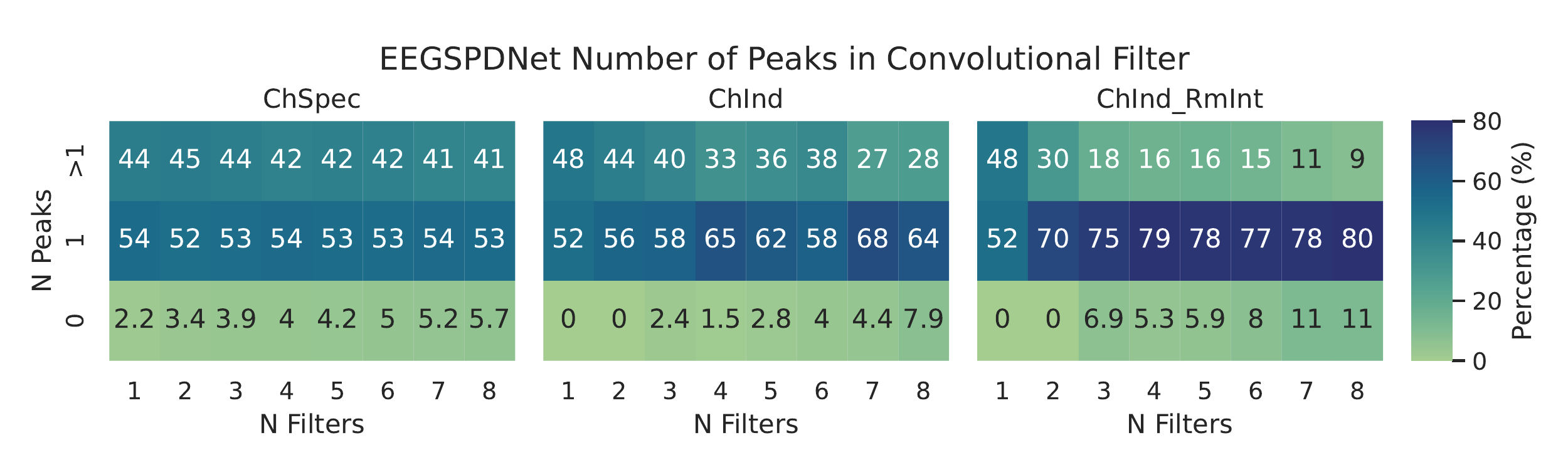}
    \caption{\color{black}
    \textbf{EEGSPDNet Multiband Occurrence via Number of Peaks in Frequency Gain}
    For every convolutional filter, the numbers of peaks in the frequency gain spectra were counted.
    Exact details of peak detection can be found in Section \ref{Sec: Methods/Analyses/Peak Detection}.
    Each subplot represents a different model sub-type, \ac{chspec} (left), \ac{chind} (center) and \ac{chind} with interband covariance removed (right).
    The x-axis shows the number of filters, $N_f$, for the given model.
    The y-axis shows the category of number of peaks found in the response of each filter (0, 1, or >1 peaks).
    The colour/annotation of each cell shows the percentage of frequency gain spectra that fell into number of peaks category for the given model (subplot) and $N_f$ (x axis).
    Therefore, any column in a heatmap sums to 100, and a value of 70 for $N_f = 2$ and N Peaks = 1 indicates that 70\% of that models frequency gain spectra had exactly 1 peak.
    Data used was all from the validation phase.
    }
    \label{Fig: Multiband}
\end{figure}

The numbers of peaks present in the frequency gain spectra were also analysed, suggesting that the networks learn "multiband" filters.
These are filters that preserve/amplify more than 1 distinct region in the frequency space.
Figure \ref{Fig: Multiband} shows that all variants of the Conv-\ac{eespdnet} learned multiband filters, with the \ac{chspec} variant having the largest proportion of multiband filters.
The electrode-frequency relevance plots provided some examples of specific multiband filters, which can be seen most prominently in Figure \ref{Fig: CovgradP5}(right hand, FFC6h) where the model has learned a filter that selected alpha/beta and high gamma specifically.

\subsection{Finding 6: Only Initial ReEig Layers were Active}
\label{Res: F6: ReEig}

\begin{figure}
    \centering
    \includegraphics[width=\textwidth]{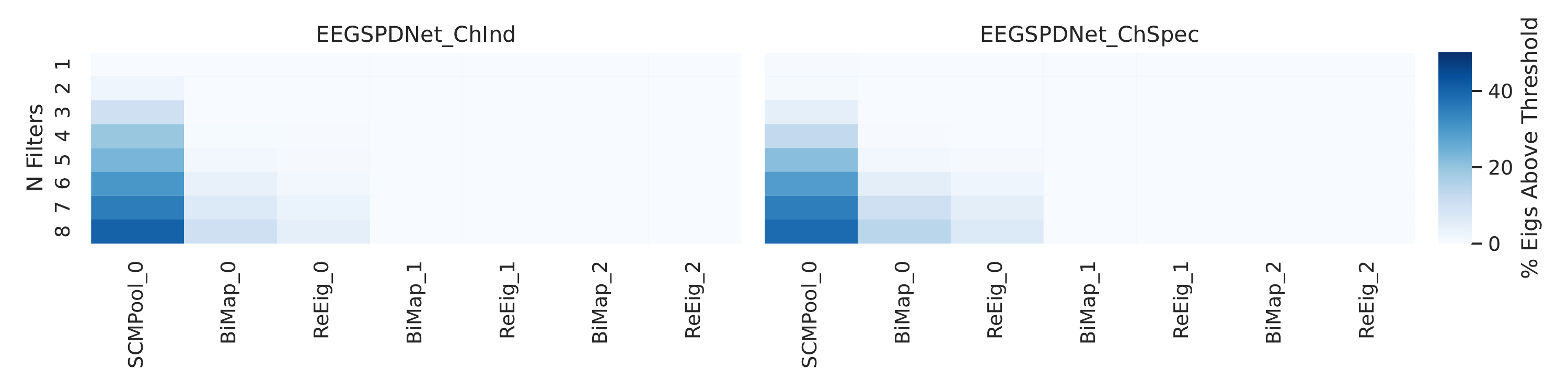}
    \caption{\color{black}
\textbf{Percentage of Eigenvalues Over ReEig Threshold.}
    Heatmaps showing proportion of eigenvalues in feature maps that are below the ReEig threshold (and would therefore be rectified).
    $N_f$ is shown on the y-axis and layer name is given on the x-axis.
    Eigenvalues were calculated after passing the data through the labelled layer.
    }
    \label{Fig: ReEig Activ}
\end{figure}

\begin{figure}
    \centering
    \includegraphics[width=\textwidth]{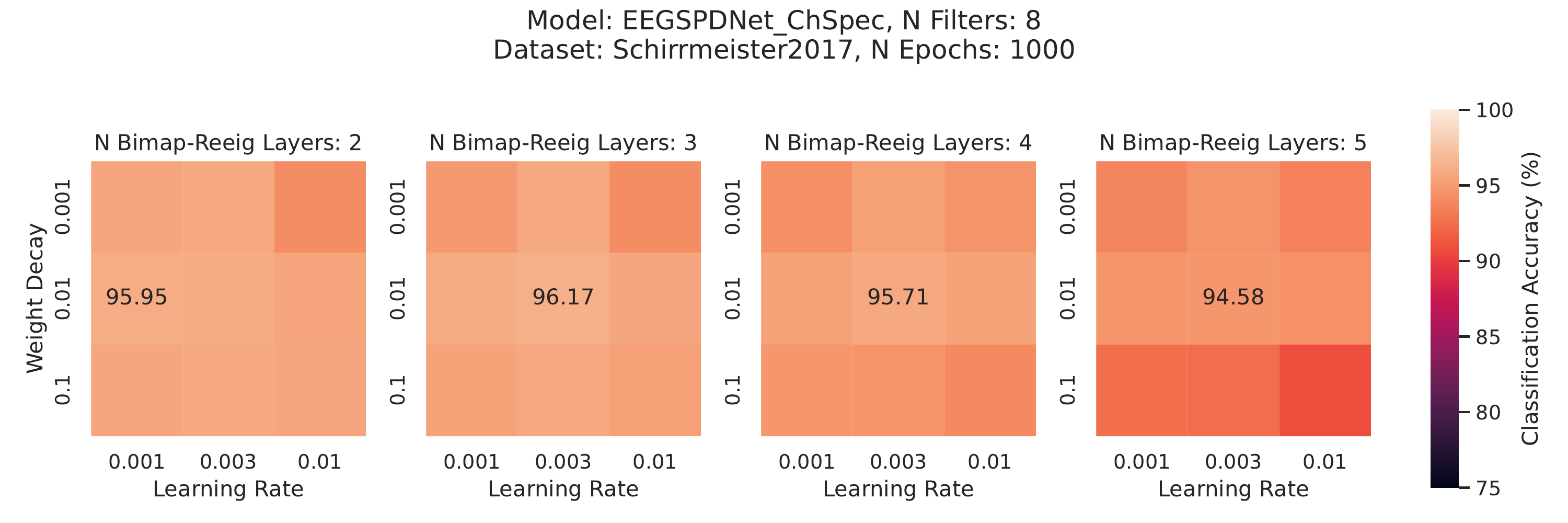}
    \caption{
    \textcolor{black}{
\textbf{Learning Rate \& Weight Decay Parameter Sweep for EEGSPNet with Different $N_{BiRe}$.}
    See Figure \ref{Fig: ParamSweep ComparisonModels} for figure structure details.
    Data shown is for \ac{eespdnet} \ac{chspec}, $N_f=8$ and $N_{BiRe} \in \{2, 3, 4, 5\}$.
    }}
    \label{Fig: ParamSweep Depth}
\end{figure}

The eigenvalues of the feature map matrices inside the some \ac{eespdnet} model were also computed.
This data is presented in Figure \ref{Fig: ReEig Activ} and shows that the majority of eigenvalues in the feature map matrices are above the ReEig activation threshold before reaching the first ReEig layer, and none remain after this first eigenvalue rectification.
In this sense, the first BiMap-ReEig pairing rectify all eigenvalues for the rest of the network.
It can also be observed that network width increased the presence of below-threshold eigenvalues.
Despite this, the initial parameter sweeps (Figure \ref{Fig: ParamSweep Depth}) still indicated that a network with a 3 pairs of BiMap-ReEig was ideal.

\subsection{Finding 7: For Filterbank Optimisation, Classifier Choice was more Important than Riemannian Metric}
\label{Res: F7: FB Opt}

\begin{figure}[ht]
    \centering
    \includegraphics[width=\textwidth]{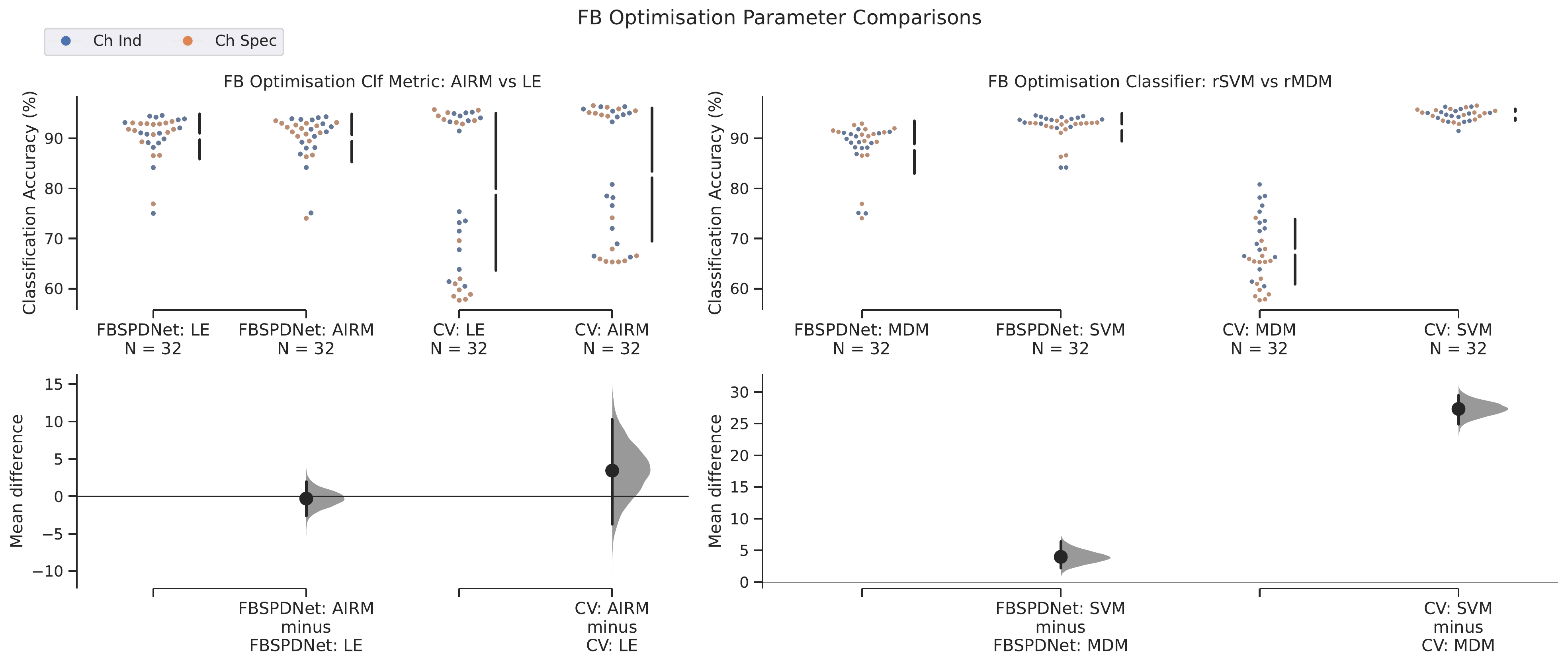}
    \caption{\color{black}
    \textbf{Filterbank Optimisation Estimation Plots}
Gardner-Altman Estimation plots showing classification results for the CV stage during filterbank optimisation, and also for final \ac{fbspdnet} classification.
Both classifiers (rMDM and rSVM) are shown, as well as both Riemannian metrics (LEM and AIRM).
For details regarding estimation plots see Figure \ref{Fig: EvalSet_EstPlot_SigMatrix}, and also \citet{hoMovingValuesData2019}.
    }
    \label{Fig: FB Opt EstPlots}
\end{figure}

The estimation plots in Figure \ref{Fig: FB Opt EstPlots} compare the effect of parameter choices (metric and classifier) at the FB optimisation stage on both the final SPDNet performance and the CV performance of the classifier used for optimisation.
The datasets used during CV and for the SPDNet are different (see Section \ref{Sec: Methods/Procedure}) and so direct comparisons between the \ac{rsvm}/\ac{rmdm} and the SPDNet are not warranted.
From this data we can see that the choice of Riemannian metric (LE or AIRM) was essentially irrelevant for final SPDNet classifier performance, even though the \ac{airm} produced marginally better results during the \ac{fbopt} stage
However, the choice of classifier (rMDM or rSVM) had a much larger impact.
The rSVM produced much higher accuracies at the \ac{fbopt} stage and while the magnitude of this effect was not replicated with the ensuing SPDNet, there was still a large difference between the two.

\subsection{Finding 8: Networks Did Not Utilise All Internal Task-Relevant Information}
\label{Res: F8: LBL}

\begin{figure}[ht]
    \centering
    \includegraphics[width=\textwidth]{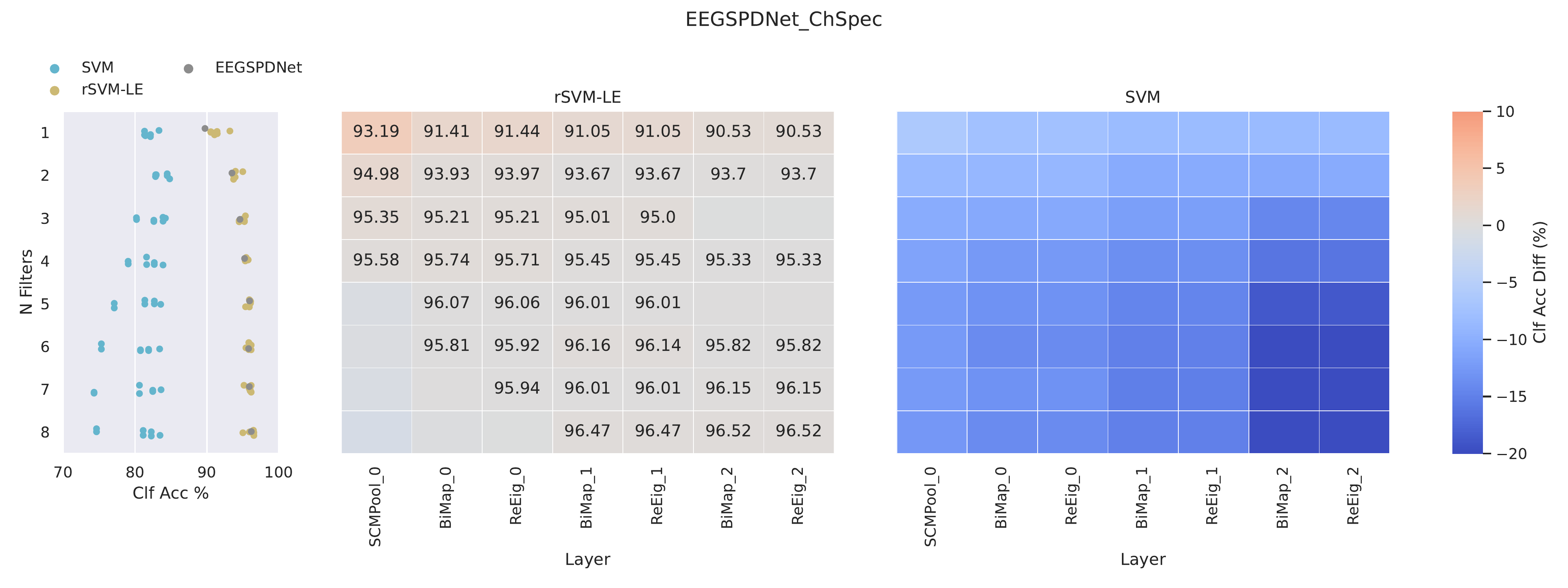}
    \caption{
    \textbf{\ac{lbl} Performance of the Channel Specific \ac{eespdnet}.}
    The left subplot column contains swarm plots showing the final accuracy of the associated \ac{eespdnet} (grey dots) and the absolute performance of the \ac{rsvm} (blue dots) and SVM (yellow dots), separated by number of filters.
    The heatmaps (centre and right subplots) show the accuracy difference of the network to an \ac{rsvm} or regular (euclidean) SVM, respectively.
    The red/blue shading of each cell shows accuracy difference between the \ac{rsvm}/SVM and the \ac{eespdnet} (\ac{rsvm}/SVM score minus \ac{eespdnet} score).
    In the cases where the \ac{rsvm} or SVM outperformed its associated \ac{eespdnet}, its absolute accuracy is annotated in the cell.
    Analysis was performed using data from the validation phase.
    }
    \label{Fig:LBL EE-SPDNet Spec}
\end{figure}

\begin{figure}[h!]
    \centering
    \includegraphics[width=\textwidth]{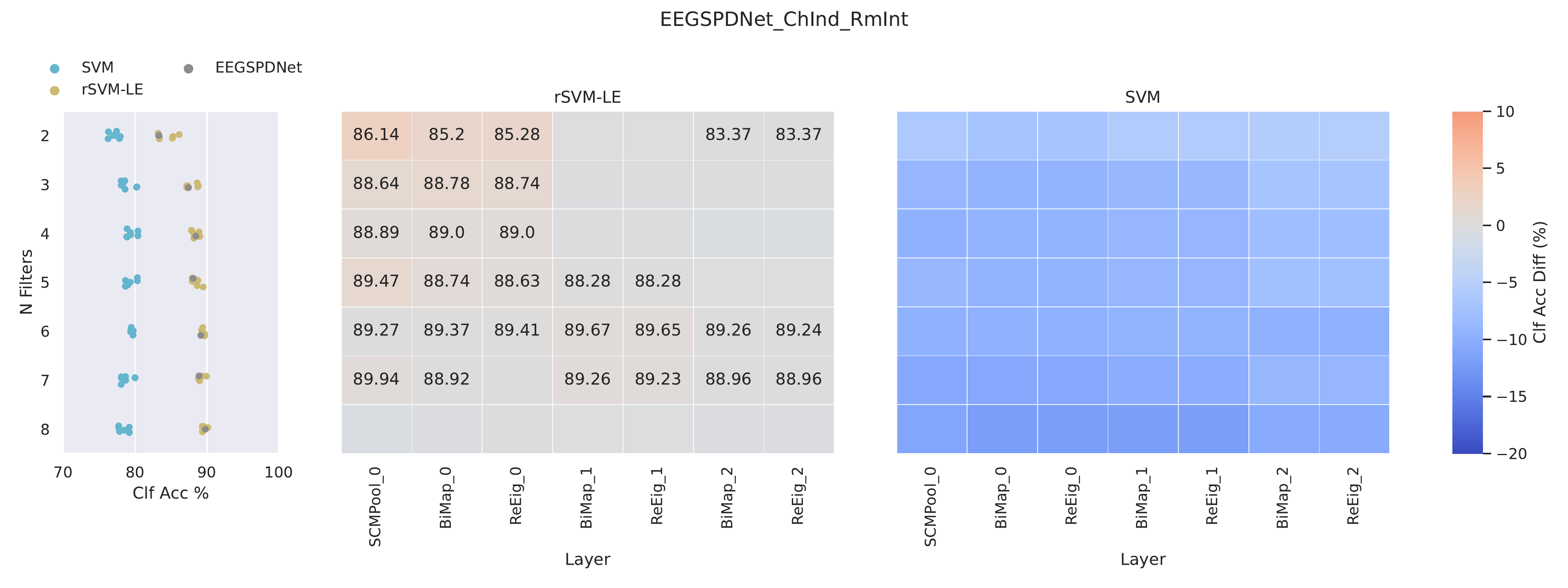}
    \caption{
    \textbf{\ac{lbl} Performance for the Channel Independent \ac{eespdnet}, without Interband Covariance.}
    \textcolor{black}{A general description of the LBL plots can be found in the caption for Figure \ref{Fig:LBL EE-SPDNet Spec}.
    }}
    \label{Fig:LBL RMINT EE-SPDNet IND}
\end{figure}

\begin{figure}[h!]
    \centering
    \includegraphics[width=\textwidth]{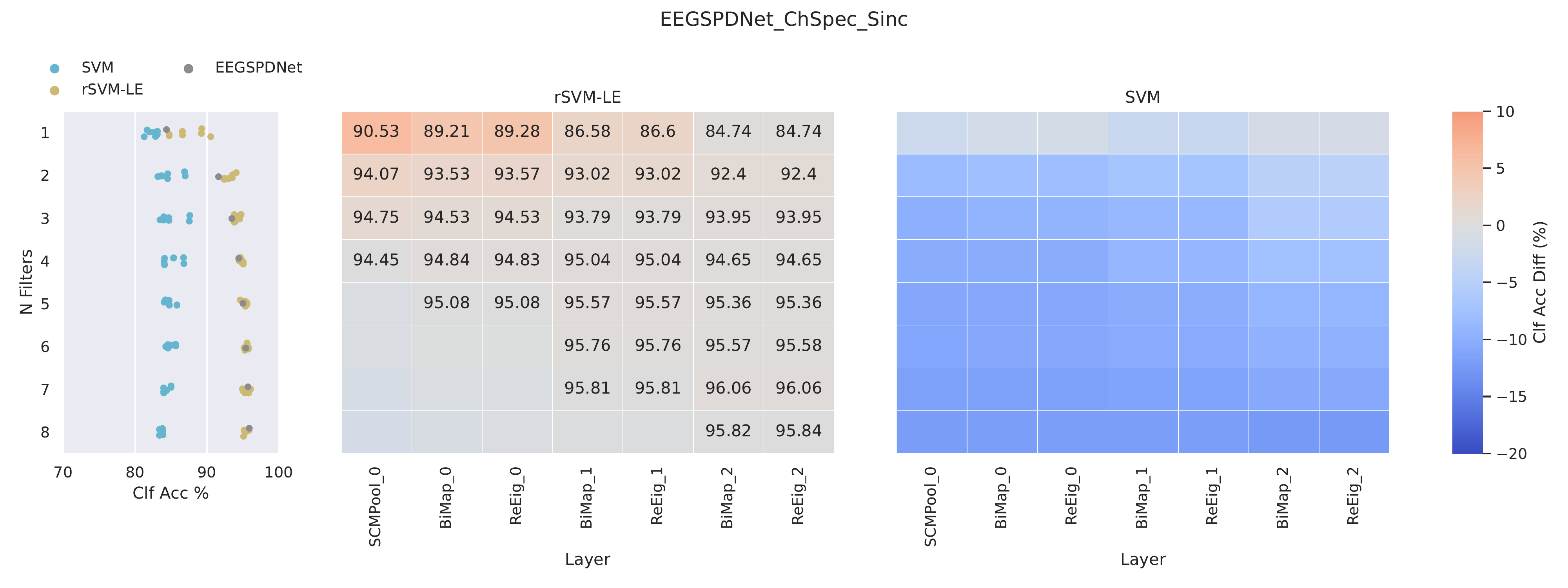}
    \caption{\color{black}
    \textbf{\ac{lbl} Performance of the Channel Specific Sinc-\ac{eespdnet}.}
    A general description of the LBL plots can be found in the caption for Figure \ref{Fig:LBL EE-SPDNet Spec}.
    }
    \label{Fig:LBL Sinc Spec}
\end{figure}

\begin{figure}[h!]
    \centering
    \includegraphics[width=\textwidth]{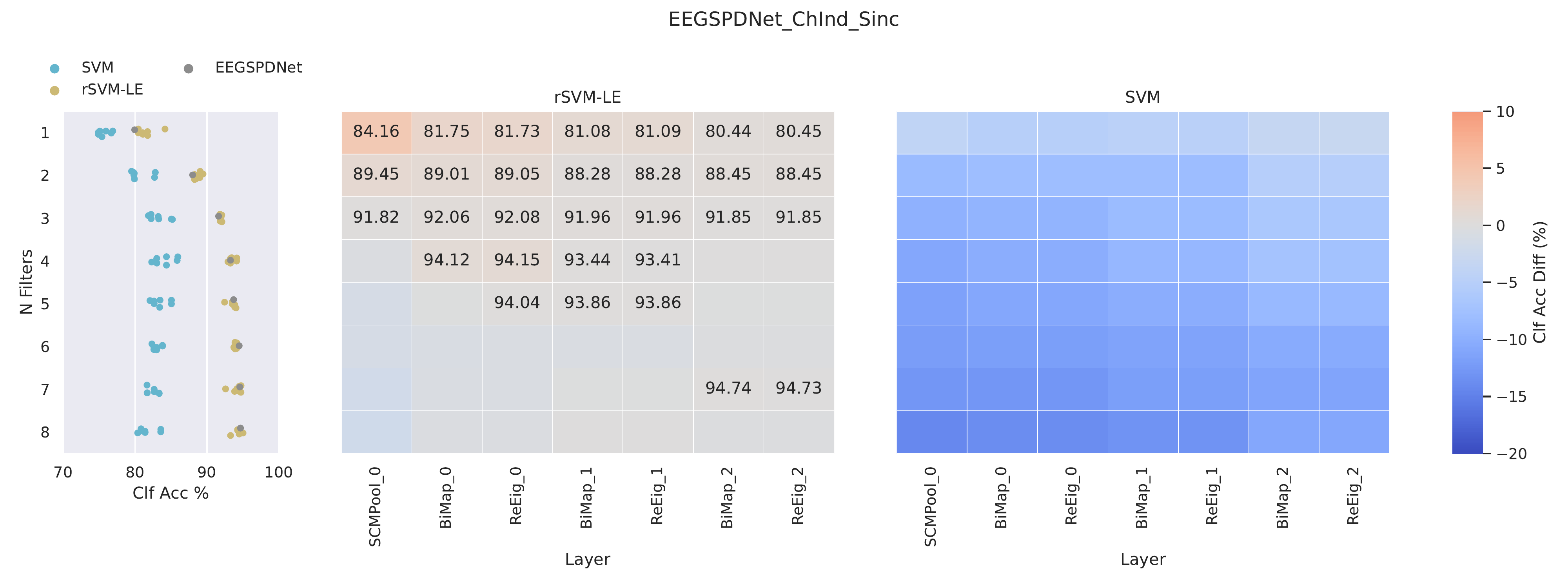}
    \caption{\color{black}
    \textbf{\ac{lbl} Performance for the Channel Independent Sinc \ac{eespdnet}.}
    A general description of the LBL plots can be found in the caption for Figure \ref{Fig:LBL EE-SPDNet Spec}.
    }
    \label{Fig:LBL Sinc Ind}
\end{figure}

The layer-by-layer classification analysis (see Figure \ref{Fig:LBL EE-SPDNet Spec}) revealed that the networks in general failed to sufficiently utilise all task-relevant information.
At at least one layer (sometimes all), every model had it's final layer classification performance beaten by an rSVM applied on that particular layer's feature maps of matrices.
Generally these accuracy differences are less than 1\%, although some (particularly at lower network width) are more stark.
Additionally, there is contrasting behaviour regarding the performance of the regular (euclidean) SVM in this analysis.
The SVM never outperformed the final layer SPDNet classification, but the patterns of underperformance do vary between model types.
Looking at the \ac{eespdnet} \ac{chspec} (\ref{Fig:LBL EE-SPDNet Spec}), the performance of the SVM was particularly bad in the final layers of the network.
This trend is more prominent for wider networks.
However, this trend is in contrast to the other network types (see Figures \ref{Fig:LBL Sinc Ind}, \ref{Fig:LBL EE-SPDNet Ind}, \ref{Fig:LBL Sinc Spec} and \ref{Fig:LBL RMINT EE-SPDNet IND}), where SVM performance is either fairly constant, or improves throughout the network.
Furthermore, when looking at $N_f = 1$, the SVM performed noticeably worse on the feature maps of the \ac{eespdnet} \ac{chspec}, something which is much less prominent for the other \ac{eespdnet} models.

\subsection{Visualisations}
\label{Res: Viz}

\subsubsection{tSNE}
\label{Res: Viz/tSNE}
\begin{figure}[ht]
    \centering
    \includegraphics[width=\textwidth]{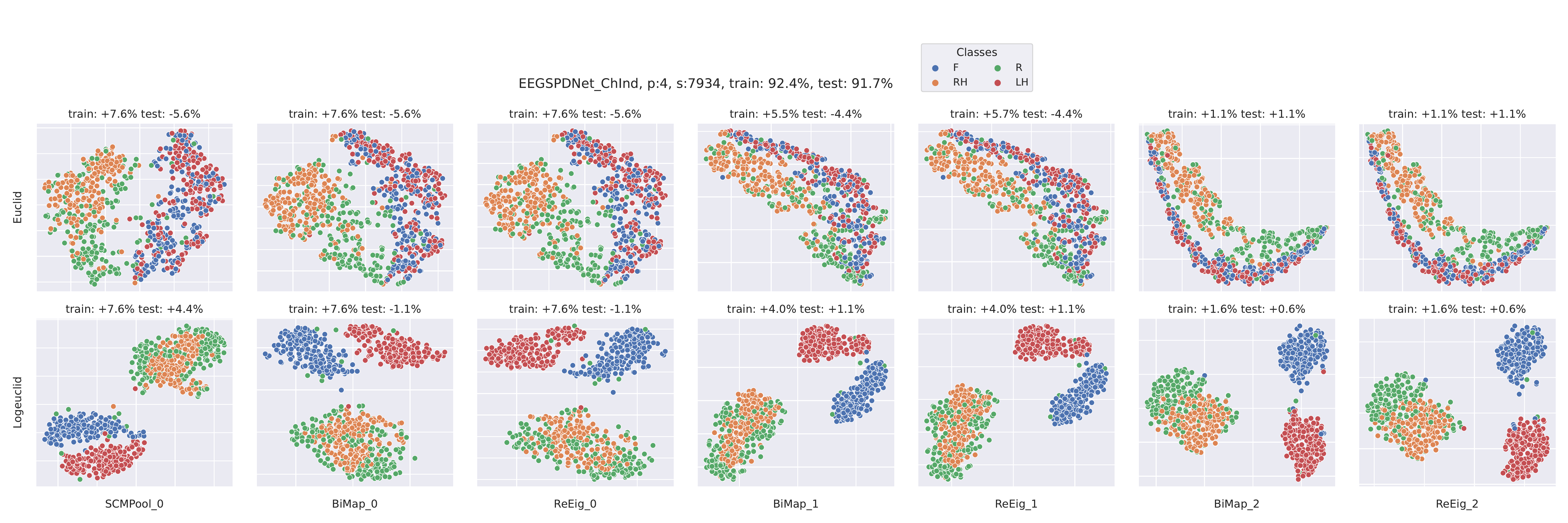}
    \caption{
    \textbf{tSNE Visualisation of Data in \textcolor{black}{Channel Independent} \ac{eespdnet}}
    Data is visualised as it passes through the layers of a \ac{chspec} \ac{eespdnet}.
    Each column shows the data after it has passed through the associated layer (subplot title).
    The top row used the Euclidean distance between matrices as the metric, the bottom used a Riemannian metric (\ac{logeuc}).
    The final training and testing accuracies of the data are given in the Figure title.
    The data plotted is test-set data (although from the train/test split of the validation set).
    The accuracy shown in each subplot title is the relative accuracy of an SVM applied to the data in the subplot.
    A positive accuracy implies that the SVM outperformed the final \ac{eespdnet} accuracy.
    \textcolor{black}{For the top row, a Euclidean SVM was used, and the bottom row used a Riemannian SVM (Log-Euclidean).}
    The colour of each point in the subplots denotes its class label (right hand, left hand, rest or feet) which is also shown in the figure legend.
    The data shown is that from a single participant for a single model, namely participant \textcolor{black}{4} from a \textcolor{black}{$N_f = 1$} channel \textcolor{black}{independent} model.
    }
    \label{Fig: TSNE Ind}
\end{figure}

\begin{figure}[ht]
    \centering
    \includegraphics[width=\textwidth]{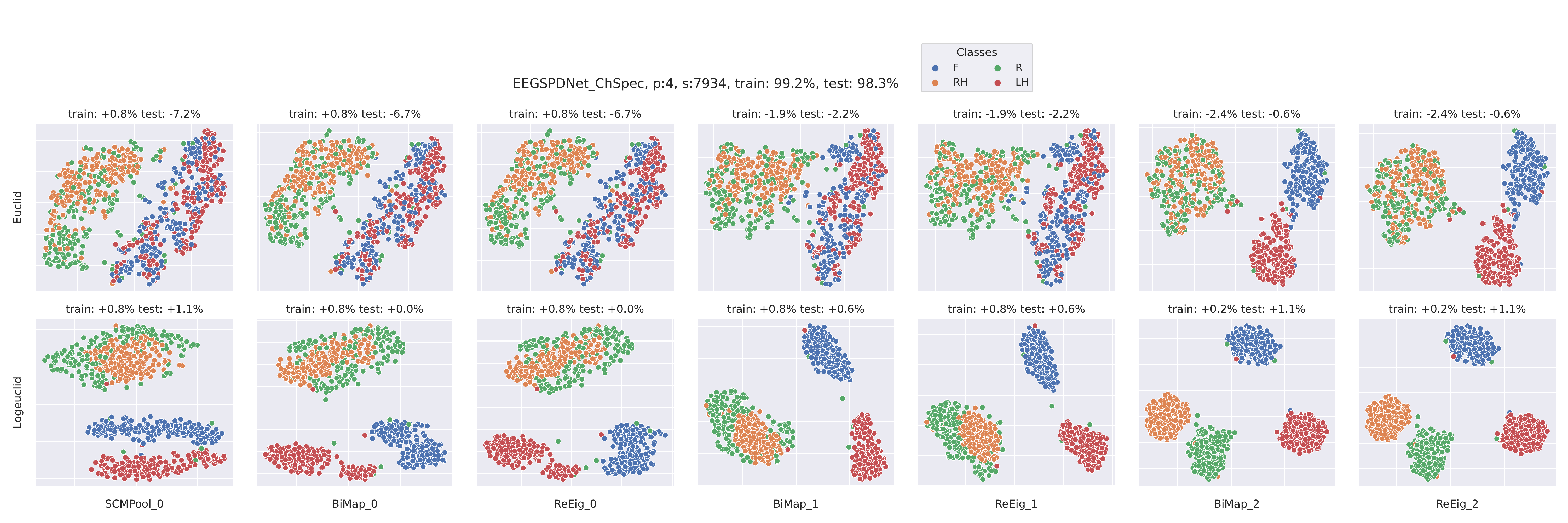}
    \caption{\color{black}
    \textbf{tSNE Visualisation of Data in Channel Specific \ac{eespdnet}}
    See Figure \ref{Fig: TSNE Ind} for more details.
    The data shown is that from a single participant for a single model, namely participant 4 from a $N_f = 1$ channel specific model.
    }
    \label{Fig: TSNE Spec}
\end{figure}

\textcolor{black}{We also visualised the \ac{lbl} data using tSNE,} two  \textcolor{black}{example}s \textcolor{black}{of this} are \textcolor{black}{shown in Figure}s \ref{Fig: TSNE Ind} \& \ref{Fig: TSNE Spec}, both are $N_f=1$ \ac{eespdnet} models.
We have chosen a high-performing participant and seed, and shown the \ac{chspec} (Figure \ref{Fig: TSNE Spec}) and \ac{chind} (Figure \ref{Fig: TSNE Ind}) variants.

Firstly, we can see that the participant-seed data shown in Figure \ref{Fig: TSNE Spec} is somewhat representative of the average accuracy trends shown in the \ac{lbl} analysis in Figure \ref{Fig:LBL EE-SPDNet Spec}.
The \ac{rsvm} classification accuracies (lower row of Figure \ref{Fig: TSNE Spec}) are at a minimum equal to the final SPDNet score, and many of the layers outperform the final layer accuracies.
In this particular example the highest outperformance occurs at the first layer and last two, instead of just the first layer as in the \ac{lbl}.
The \ac{svm} scores stay below final layer classification, but do improve through the network especially after the second BiMap layer.
This does not exactly match the overall trend seen in the \ac{lbl} where Euclidean accuracies stay roughly constant.

\begin{figure}[h!]
    \centering
    \includegraphics[width=\textwidth]{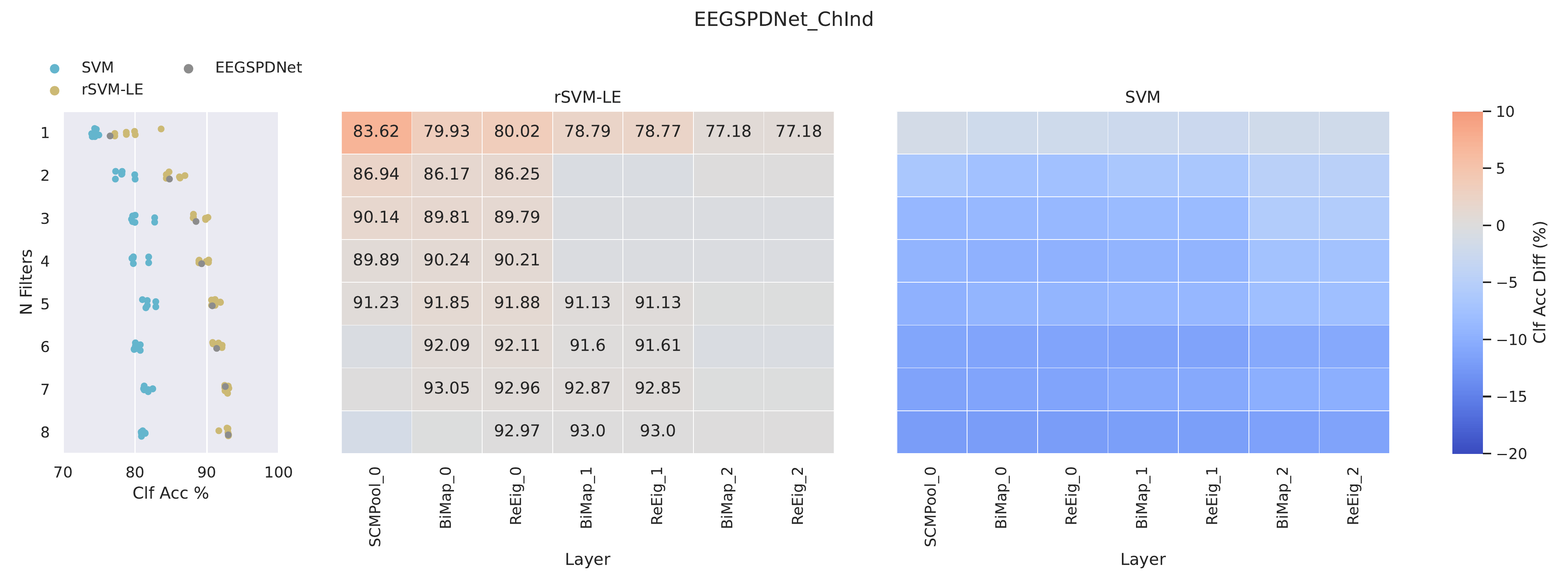}
    \caption{
    \textbf{\ac{lbl} Performance of the Channel Independent \ac{eespdnet}.}
    \textcolor{black}{A general description of the LBL plots can be found in the caption for Figure \ref{Fig:LBL EE-SPDNet Spec}.
    }}
    \label{Fig:LBL EE-SPDNet Ind}
\end{figure}

In the \ac{chind} setting, the tSNE data in Figure \ref{Fig: TSNE Ind} with the \ac{rsvm} matches the trend seen in the \ac{lbl} data in Figure \ref{Fig:LBL EE-SPDNet Ind}.
The \ac{rsvm} outperforms the \ac{eespdnet} strongly early in the  network with the effect decreasing as the data is passed through the layers.
This is partially observable in the tSNE data itself, as the visual separation between classes barely changes through each layer.
The \ac{svm} scores in Figure \ref{Fig: TSNE Ind} are slightly atypical in that the final two layers have outperformed the final layer of the SPDNet.
The \ac{lbl} data suggests that in general the \ac{svm} does not perform better than the final layer of the SPDNet, and that this does not change much as the data is propagated through the network.

When comparing the two tSNE Figures, we are comparing the \ac{chind} (Figure \ref{Fig: TSNE Ind}) filtering to \ac{chspec} (Figure \ref{Fig: TSNE Spec}), as all other variables (participant, seed etc) are held constant.
It is difficult to draw solid conclusions from the tSNE visualisations, however it is interesting to note how the visual separation of classes varies between these two Figures.
The \ac{chspec} model succeeds in separating classes that are not visually separated in the \ac{chind} variant, but only in the final two layers.
By the end of the \ac{chspec} network, all four classes a visually separated for the Riemannian tSNE, whereas "Rest" and "Right Hand" are still mixed for the \ac{chind} at the same point in the network.
A similar pattern occurs for the Euclidean tSNE, except with "Feet" and "Left Hand".

\color{black}
\color{black}

\subsubsection{Electrode-Frequency Relevance}

\begin{figure}[ht!]
    \centering
    \includegraphics[width=\textwidth]{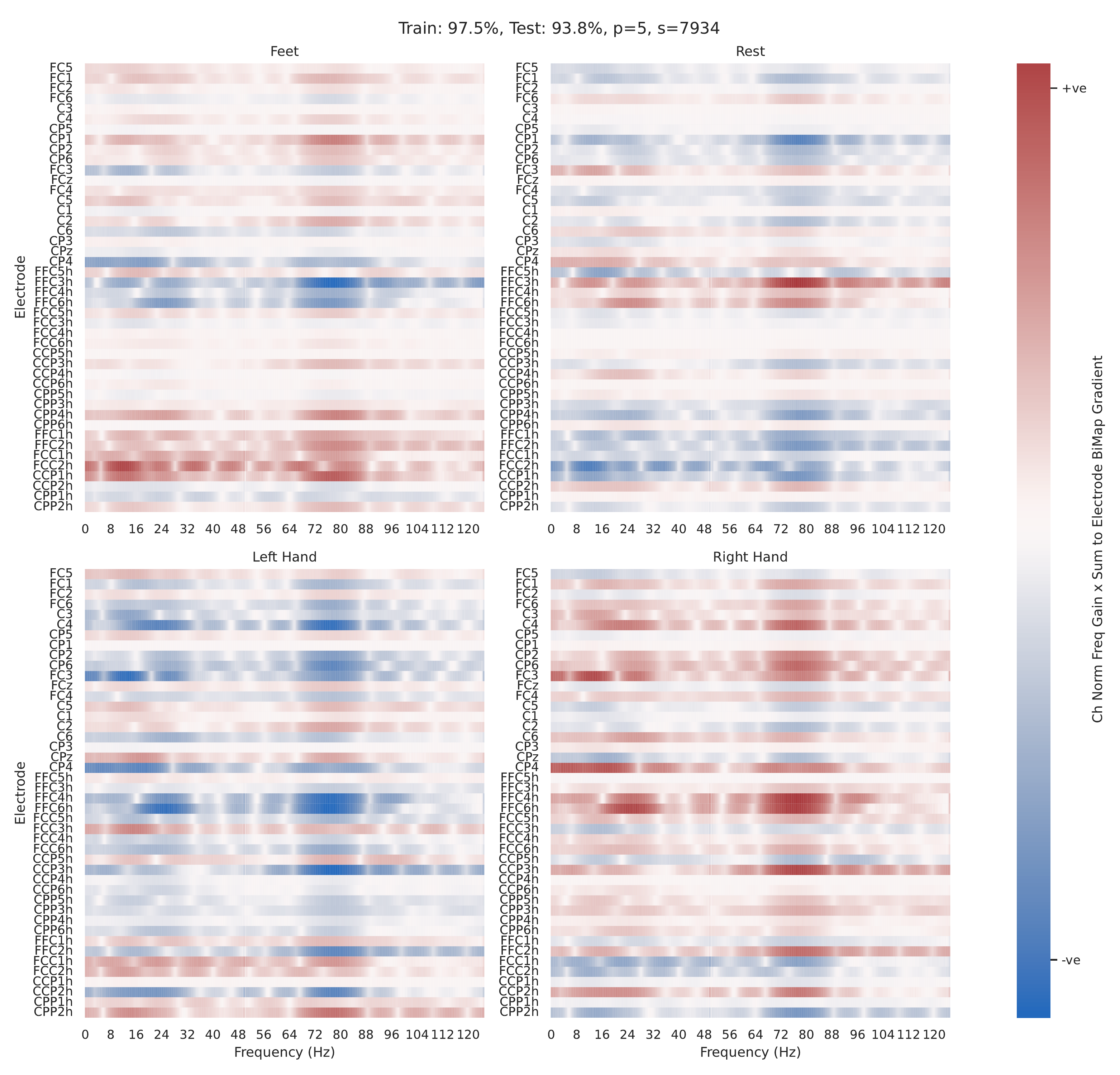}
    \caption{\color{black}
    \textbf{Electrode Frequency Relevance Plots for Participant 5, Schirrmeister2017.}
    Per class heatmaps, that show the product of normalised frequency gain spectra with aggregated-to-electrode gradient values.
    Precise details regarding calculations can be found in Section \ref{Sec: Methods/Analyses/Electrode Relevance}.
    Each subplot shows each class, with frequency on the x-axis, and electrode number on the y-axis.
    Hue indicates relevance i.e. a change in that frequency, at that electrode results in a positive or negative increase in the softmax prediction for that class.
    Data is shown for a single, fully-trained \ac{eespdnet} model (one participant, one seed), details of which can be seen in the figure title.
    The model used \ac{chspec} filtering and had $N_f$
    }
    \label{Fig: CovgradP5}
\end{figure}

\begin{figure}[ht!]
    \centering
    \includegraphics[width=\textwidth]{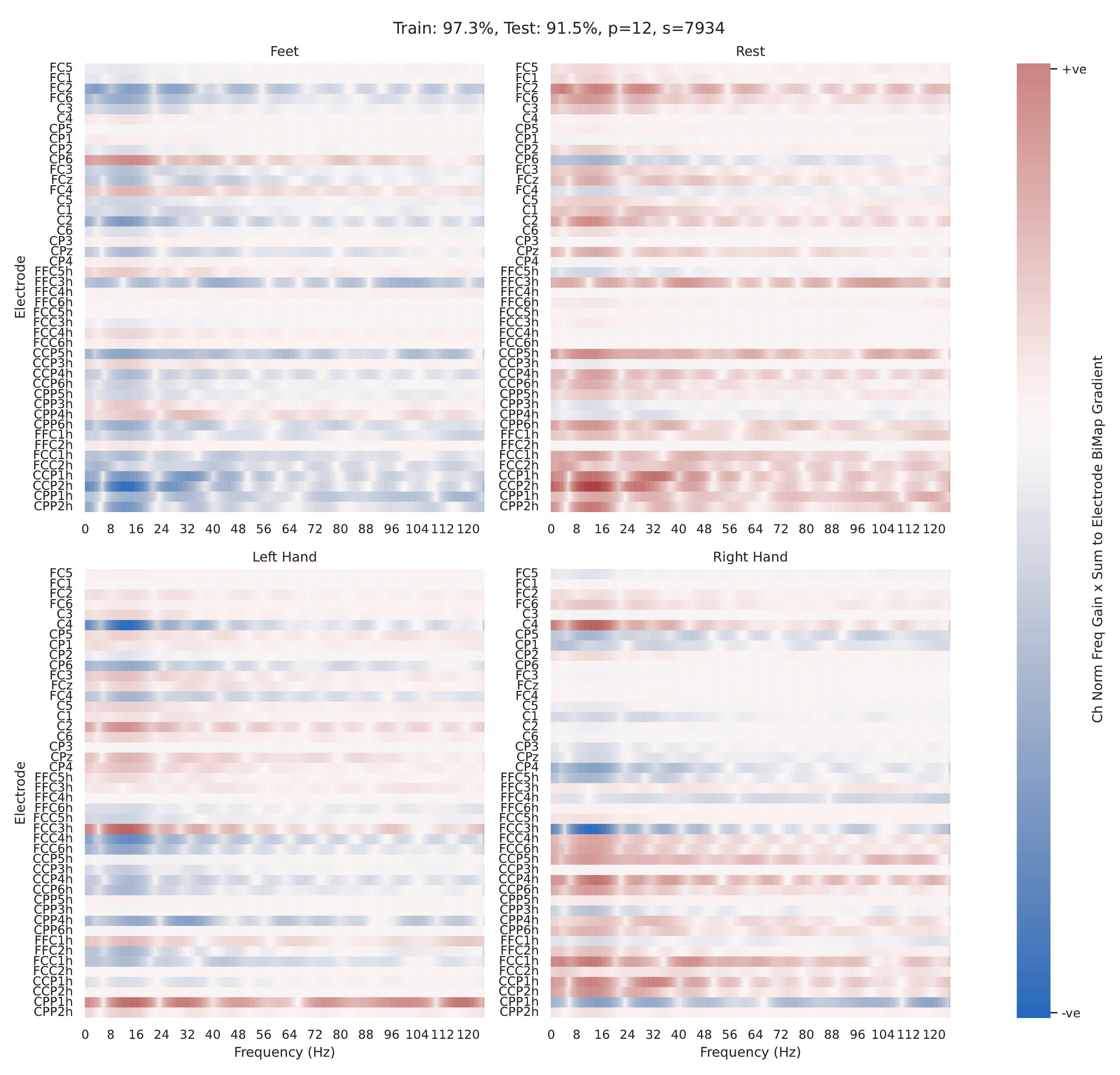}
    \caption{\color{black}
    \textbf{Electrode Frequency Relevance Plots for Participant 12, Schirrmeister2017.}
    see Figure \ref{Fig: CovgradP5} for plot details.
    The model used \ac{chspec} filtering and had $N_f=1$
    }
    \label{Fig: CovgradP12}
\end{figure}

Figures \ref{Fig: CovgradP5} and \ref{Fig: CovgradP12} show two electrode frequency relevance plots for \ac{chspec} \ac{eespdnet}, $N_f=1$.
These two runs were chosen as they highlight a number of interesting features.
Firstly, a number of patterns emerge which are typical for subject-specific motor EEG data such as inverted left and right side usage and subject specific spatial and frequency usage. 
In particular, we note the usage of the high-gamma region, which is dominant in Figure\ref{Fig: CovgradP5}, and absent in Figure \ref{Fig: CovgradP12}.
Secondly, Figure \ref{Fig: CovgradP5} also shows a strong example of the multiband filter (Electrode FFC6h) selecting two distinct and physiologically plausible frequency region.

\subsubsection{BiMap Gain}

\begin{figure}[ht]
    \centering
    \includegraphics[width=\textwidth]{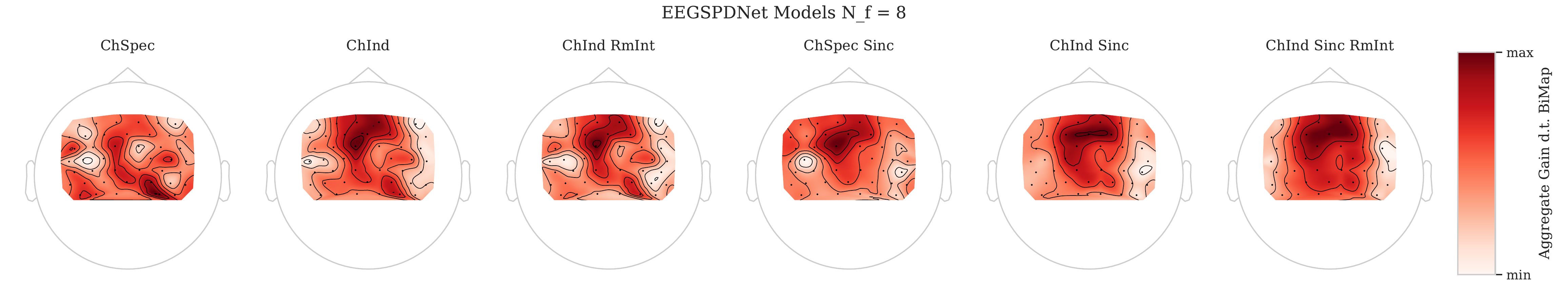}
    \caption{\color{black}
    \textbf{BiMap Gain for $N_f=8$ EEGSPD Models}
    Head plots showing aggregate electrode covariance gain from the first BiMap layer of trained models.
    For an input covariance matrix, an equally sized gain matrix can be computed which shows the contribution of each element in the covariance matrix to the output matrix.
    This symmetric can be row (or column) summed to get the contribution of an individual electrode (via it's variance and covariance pairs).
    See Section \ref{Sec: Methods/Analyses/Bimap Gain} for more details.
    The colourbar min-max values were set per plot, meaning that absolute comparisons between models is not possible.
    }
    \label{Fig: BiMap Gain NF=8}
\end{figure}

In Figure \ref{Fig: BiMap Gain NF=8} we visualise the gain applied to covariance matrix elements as topographical head plots. 
These plots highlight the similarity in spatial gain between models with and without interband covariance, and also the spatial dissimilarity of the \ac{chspec} model.
Supplementary Materials Figure \ref{Fig: BiMap Gain Corr AND Sinc vs Conv Scatter} (left) shows the correlation between these topographical headplots, quantifying the above statements.

\color{black}

\section{Discussion}
\label{Discussion}

\subsection{\acfp{drn} versus ConvNets}
\label{Disc: Main Accs}

The results presented in Figure \ref{Fig: EvalSet_EstPlot_SigMatrix} and Section \ref{Res: F1: Main Accs} strongly suggest that \textcolor{black}{\acp{drn}} can outperform the convolutional neural networks \textcolor{black}{on motor-EEG datasets.
The only ConvNet model to outperform a \ac{drn} was ShallowFBCSPNet, and it was still beaten by \ac{eespdnet} and TSMNet.}
In particular, the 8 \textcolor{black}{filter} channel specific \textcolor{black}{Conv-}\ac{eespdnet} has the highest accuracy \textcolor{black}{and most significant results, and from a pure decoding ability perspective,} and could be seen as the "best" of the models presented in this study.
\color{black}

\subsection{Hyperparameter and Network Design Choices and their Influence on Behaviour}
\label{Disc: Hyperparams}
\subsubsection{Filter Type}
\label{Disc: Filter Type}
This study employed two types of learned filterbanks: \ac{fbspdnet} and Sinc-\ac{eespdnet} used learnable sinc convolutions, while Conv-\ac{eespdnet} utilized standard convolutional filters. 
The latter can develop into more complex forms, such as dual-bandpass or multibandpass filters, allowing multiple frequency bands to pass through.
Our analysis of multibandpass filter occurrence aimed to investigate performance disparities between Conv-\ac{eespdnet} and Sinc-\ac{eespdnet}, which aside from filtering style, are identical.
While other complex filter types may exist, they were not explored in this study.

Manual examination of frequency gain spectra (e.g., Figure \ref{Fig: CovgradP5}, electrode FFC6h) and a peak detection algorithm (Section \ref{Sec: Methods/Analyses/Peak Detection}) revealed the presence of multiband filters. 
Results are presented in Figure \ref{Fig: Multiband} and Finding 5 (Section \ref{Res: F5: Multiband}). 
Note that the peak detection algorithm's parameters were developed heuristically, potentially affecting the accuracy of multiband filter counts.

Several correlations between multiband filters and decoding performance can be observed in the data.
Conv-\ac{eespdnet}'s superior performance over Sinc-\ac{eespdnet} suggests an advantage in exploring beyond sinc-space filters.
Channel specific Conv-\ac{eespdnet} models exhibited more multibands and better performance than \ac{chind} models. 
The best-performing model had a high proportion and absolute number of multiband filters. 
Interestingly, while the proportion of multiband filters decreased with network width, their absolute number increased, correlating with improved decoding performance.

However, in the \ac{chind} setting, Sinc-\ac{eespdnet} outperformed Conv-\ac{eespdnet}, suggesting that the multiband filters present in Conv-\ac{eespdnet} may have degraded decoding performance, in this context.

In conclusion, multiband filters were identified in Conv-\ac{eespdnet}'s learned filterbanks, but their precise impact on decoding performance remains unclear beyond the observed correlations. 
Further investigation is warranted, with potential methodologies outlined in Section \ref{Disc: Future}.

\subsubsection{Channel Specificity}
\label{Sec: Discussion/Hyperparams/ChSpec}

\begin{figure}
    \centering
    \includegraphics[width=\textwidth]{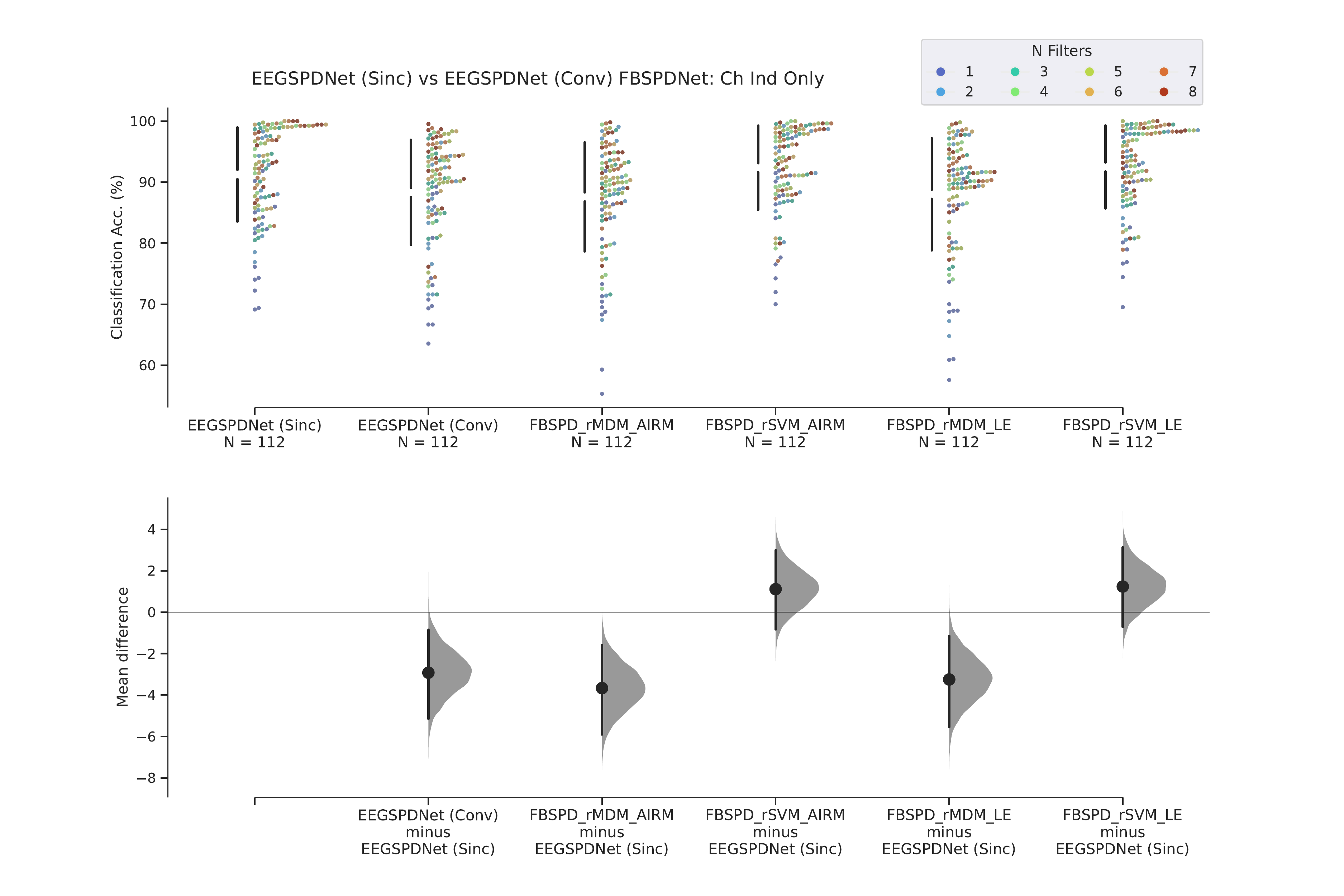}
    \caption{
    \textcolor{black}{
\textbf{Estimation Plot for Channel Independent Models.}
    A general description of estimation plot structure can be found in Figure \ref{Fig: EvalSet_EstPlot_SigMatrix}.
    Here we show all \ac{chind} \ac{eespdnet} and \ac{fbspdnet} variants from the hyperparameter optimisation phase on Schirrmeister2017.
    }}
    \label{Fig: ChInd EstPlot}
\end{figure}

\begin{figure}
    \centering
    \includegraphics[width=\textwidth]{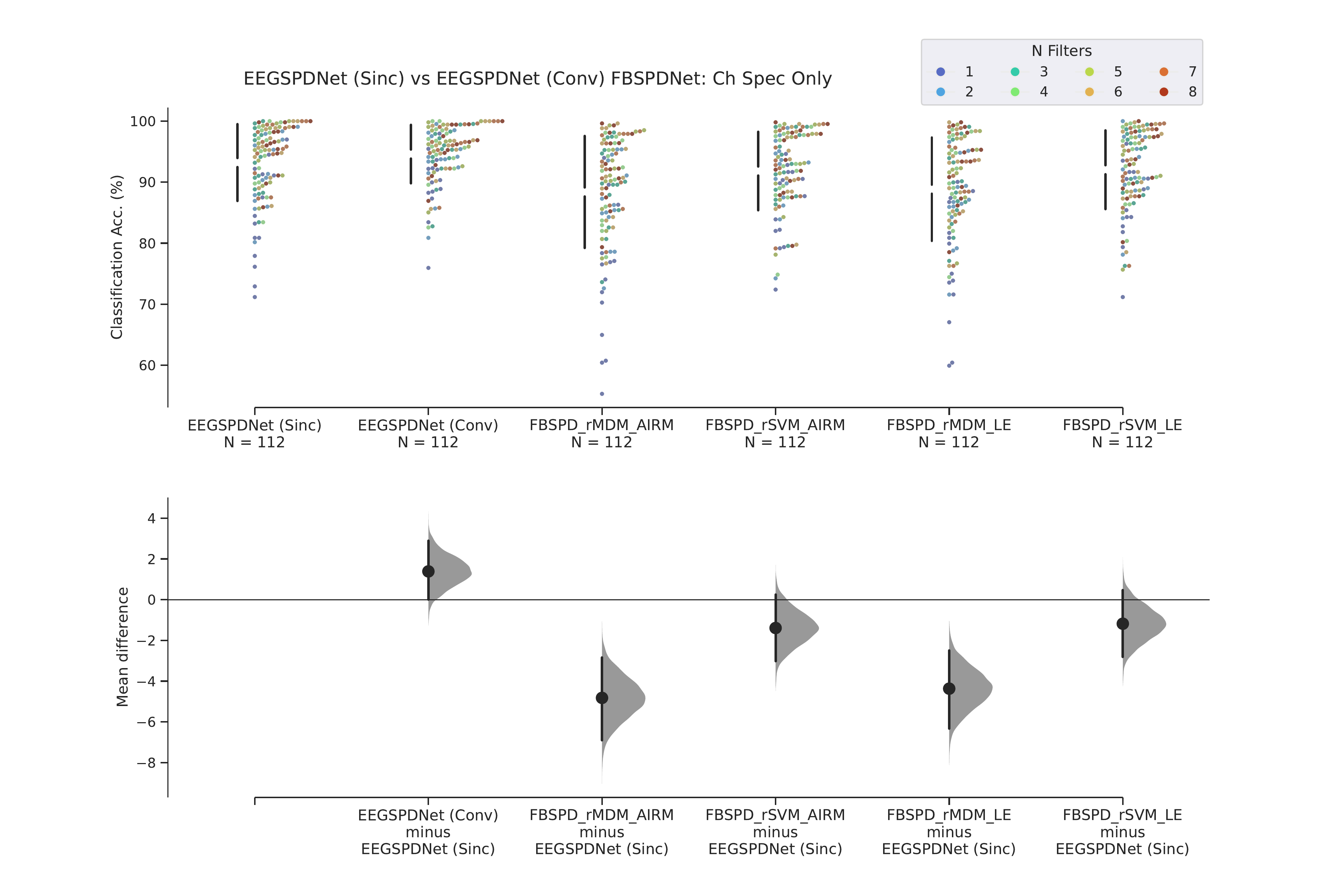}
    \caption{\textcolor{black}{
\textbf{Estimation Plot for Channel Specific Models.}
    A general description of estimation plot structure can be found in Figure \ref{Fig: EvalSet_EstPlot_SigMatrix}.
    Here we show all \ac{chspec} \ac{eespdnet} and \ac{fbspdnet} variants from the hyperparameter optimisation phase on Schirrmeister2017.
    }}
    \label{Fig: ChSpec EstPlot}
\end{figure}

\begin{figure}
    \centering
    \includegraphics[width=\textwidth]{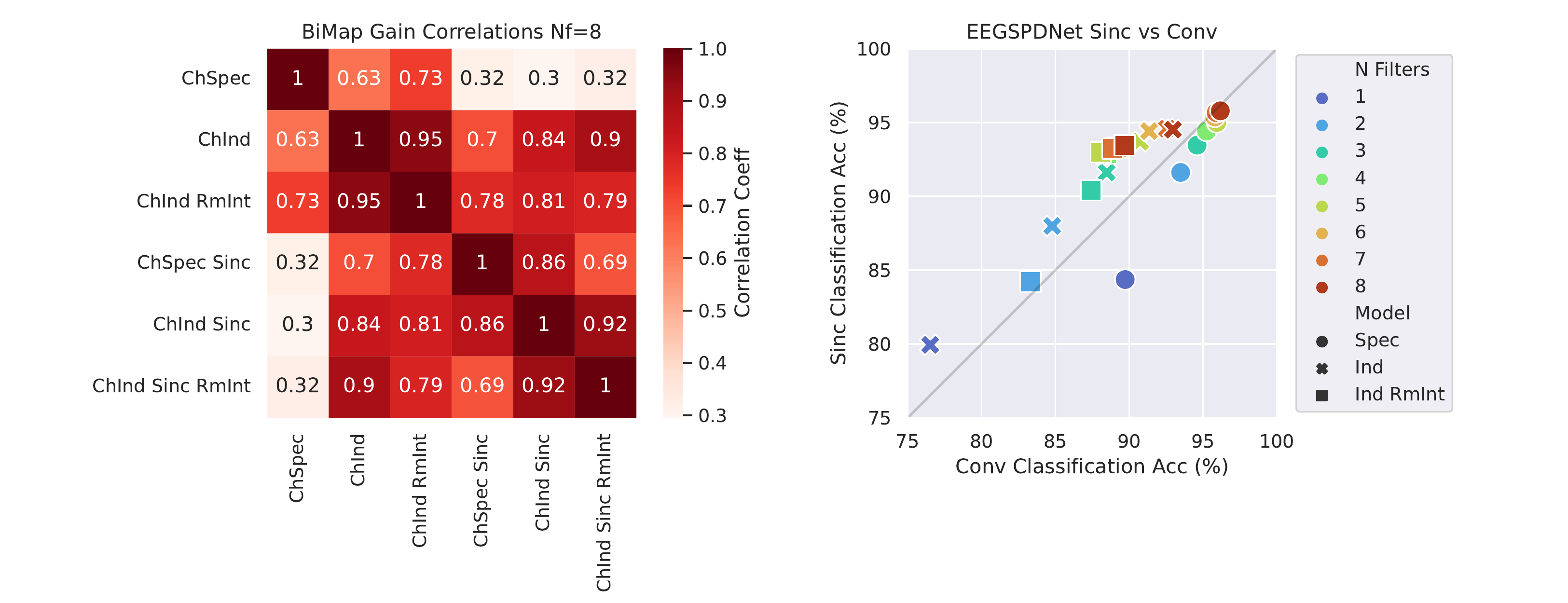}
    \caption{\textcolor{black}{
\textbf{BiMap Gain Correlations (left) and Classification Accuracy Scatterplot for Sinc and Conv \ac{eespdnet} (right).}
    Left: Correlation coefficient matrix for spatial BiMap gain values seen in Figure \ref{Fig: BiMap Gain NF=8}.
    Data highlights the similarity in spatial usage between \ac{chind} and \ac{chind} RmInt models and dissimilarity of \ac{chspec} with many other models.
    Right: A general description of classification accuracy scatter plots can be found in the caption for Figure \ref{Fig: EEvsBO_SpecVsInd_RMINT}.
    }}
    \label{Fig: BiMap Gain Corr AND Sinc vs Conv Scatter}
\end{figure}

The study's best classification performance came from an 8 filter \ac{chspec} setup paired with standard convolutional filters, outperforming \ac{chspec} sinc/bandpass filters (Findings 1 \& 2). 
Channel specific configurations increase network complexity without increasing width, allowing per-electrode frequency region selection, especially at higher network widths.
However, not every model benefited from \ac{chspec} filtering, namely \ac{fbspdnet} performed better with \ac{chind} filtering at $N_{f} \geq 2$ (Figure \ref{Fig: EEvsBO_SpecVsInd_RMINT} (center)).
In \ac{chspec} settings, Conv-\ac{eespdnet} outperformed Sinc-\ac{eespdnet}, while the opposite was true for \ac{chind} settings (Figures \ref{Fig: BiMap Gain Corr AND Sinc vs Conv Scatter} (right), \ref{Fig: ChSpec EstPlot} \& \ref{Fig: ChInd EstPlot}).

The reasons for superior performance requiring both channel specificity and convolutional filtering remain unclear.
The complex filter types available Conv-\ac{eespdnet} may be most effective at the per-electrode level, which is potentially supported by the data at $N_{f}=1$ (Figure \ref{Fig: EEvsBO_SpecVsInd_RMINT}(center)), where the largest improvement due to \ac{chspec} filtering can be observed.
This suggests only a subset of electrodes may require complex filters for optimal decoding.
The increased number of multiband filters in \ac{chspec} models could support this theory, though it likely also correlates with the overall filter count. 
While \ac{chspec} Conv-\ac{eespdnet} performed best overall, Sinc-\ac{eespdnet} and \ac{fbspdnet} outperformed it in \ac{chind} settings.
Additionally, no \ac{chind} \ac{eespdnet} model was brought forwards to the evaluation set stage, so it becomes difficult to ascertain the exact role of channel specificity in Conv based \acp{drn} for multiple datasets.

TSMNet, which uses channel independent filtering (see Table \ref{Tab: DeepRieEEG_FB}), achieved comparable accuracies to \ac{chspec} \ac{eespdnet}. 
However, TSMNet's additional features (spatial filtering and SPD batch normalization) make it difficult to isolate the impact of specific layers on decoding accuracy.

Channel independent models with deleted interband covariance showed slightly lower performance, indicating the positive role of interband covariance in classification. 
Channel specific filtering naturally includes interband covariance for models with $N_f > 1$.

\subsubsection{Filterbank Optimisation for FBSPDNet}
\label{Disc: FB Opt}

\begin{figure}
    \centering
    \includegraphics[width=0.8\textwidth]{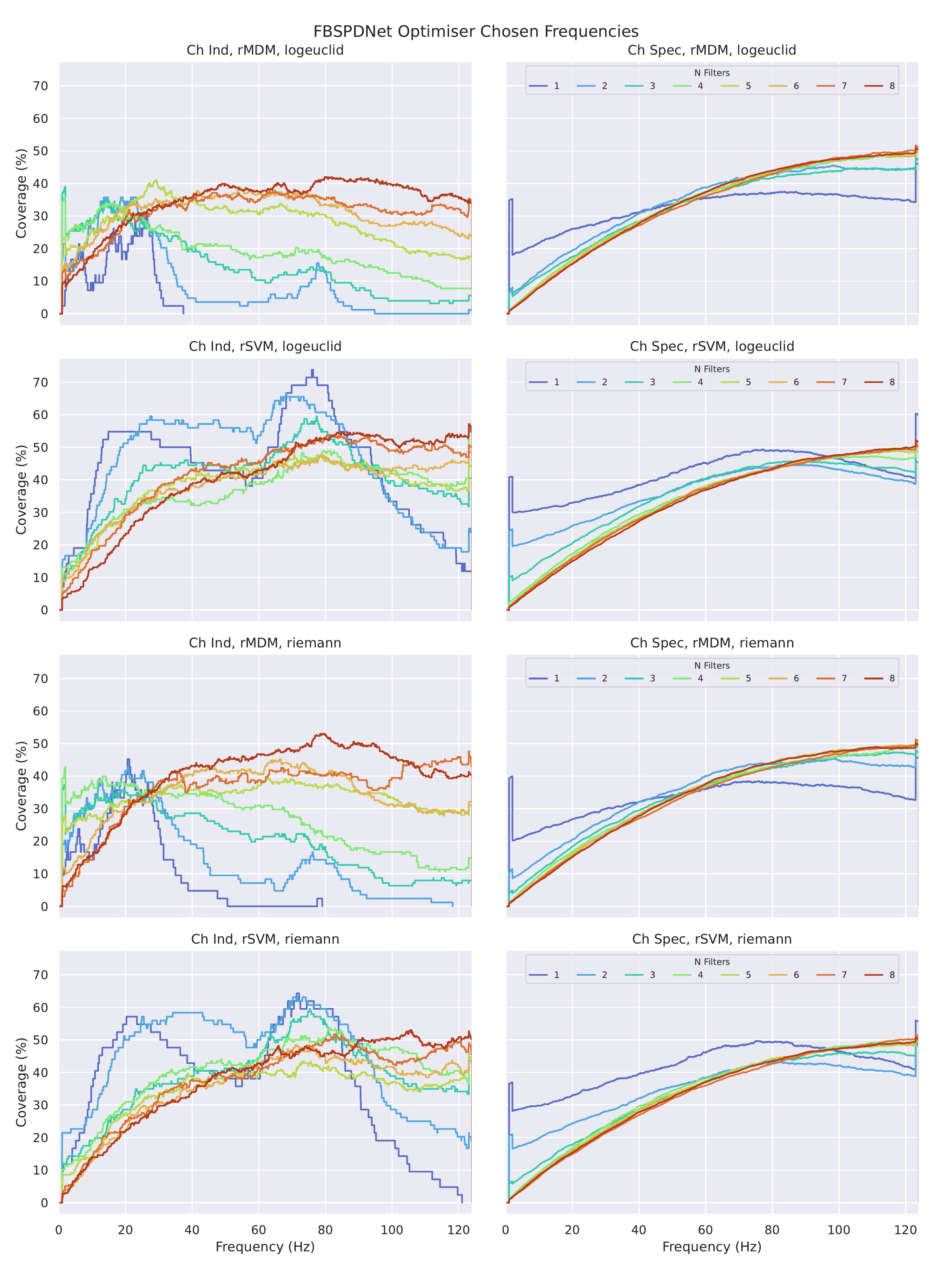}
    \caption{\color{black}
    \textbf{Frequency Band Selection Distributions all \ac{fbspdnet} variants.}
    A general description of the frequency band distribution plots can be found in the caption for Figure \ref{Fig:FBSPD ChosenFreqs}.
    Each subplot title respectively shows filter specificity, classifier, and Riemannian metric.
    }
    \label{Fig: FBSPD All Chosen Freqs}
\end{figure}

Despite FBSPDNet's modest decoding accuracies, the filterbank optimisation analysis yielded noteworthy insights.
Figure \ref{Fig: FB Opt EstPlots} and Finding 7 (Section \ref{Res: F7: FB Opt}) show that the choice of Riemannian metric (\ac{logeuc} vs \ac{airm}) minimally impacted \ac{fbspdnet} decoding and only slightly affected \ac{rsvm} decoding (\ac{airm} performed marginally better). 
This suggests that the Riemannian metric choice is not critical at the filterbank stage, implying task-relevant information for each metric likely resides in similar frequency regions.
Figure \ref{Fig: FBSPD All Chosen Freqs} (Section \ref{Sec: Appdx: Chosen Freqs}) further supports this, demonstrating that classifier choice had a more significant effect on frequency region distributions than the metric choice. 
The \ac{rsvm} consistently outperformed \ac{rmdm} in both cross-validation FB Optimisation and eventual SPDNet classification, with a larger performance gap in the CV setting.

\ac{rsvm} and \ac{rmdm} exhibited distinct frequency selection distributions. 
\ac{rsvm} favoured more high-gamma regions, especially at low $N_f$, while \ac{rmdm} generally showed lower frequency distributions, suggesting thinner band selection.
As $N_f$ increases, the filterbank's effect becomes less discernible, particularly in the \ac{chspec} setting.
This trend may indicate the optimiser's difficulty in handling numerous parameters \citep{frazierTutorialBayesianOptimization2018}, leading to semi-random selections.
This also explains the shape of the curves seen in Figure \ref{Fig: FBSPD All Chosen Freqs}, which are tending towards a random equal distribution of frequency regions (which would be expected to be a round curve centred at ~124Hz\footnote{
\color{black}
This effect stems from the bandpass filter implementation.
The optimiser adjusts a low-frequency cutoff and a bandwidth for each band. 
Cutoffs exceeding the Nyquist frequency are clamped before filter kernel computation. 
however, since the maximum bandwidth and maximum low cutoff are near the Nyquist frequency, the distribution of possible bands centres close to it.
}).
This would also explain the accuracy plateau observed at higher values of $N_f$.

\subsubsection{EEGSPDNet vs FBSPDNet}
\label{Disc: EEGSPD vs FBSPD}

EEGSPDNet outperforms \ac{fbspdnet} in both the hyperparameter selection stage and then the final evaluation across multiple datasets, suggesting that learning the filterbank via backpropagation is preferable to optimisation in a separate stage.
However, \ac{fbspdnet} offers immediate interpretability, through filterbank values and with \ac{fbopt} CV scores potentially indicating final performance. 
FBSPDNet's limitation lies in its inability to optimise in the entire convolutional space, precluding learning of complex filter types observed in Conv-\ac{eespdnet} models.

\subsubsection{Network Width}
\label{Disc: Width}

Network width, defined by the number of filters $N_f$ (see Section \ref{Sec: Methods/NetworkDesign/WidthAndChSpec}), was tested across all models. 
Increasing $N_f$ generally improved classification ability, with the largest improvement typically observed between $N_{f}=1$ and $N_f=2$.
This performance boost could be attributed to increased network complexity and over-parameterisation, potentially allowing for "lucky" sub-networks at initialisation \citep{frankleLotteryTicketHypothesis2019}.

\subsubsection{Network Depth}
\label{Disc: Depth}

Network depth, defined by the number of BiMap-ReEig pairs ($N_{BiRe}$), was set to 3 for the proposed networks.
An initial parameter sweep suggested $N_{BiRe}=3$ as optimal, though this result should be interpreted cautiously due to the coarse nature of the sweep and the constant relative size of the following layer.

Only the first ReEig layer was found to be active in trained networks (Finding 6, Section \ref{Res: F6: ReEig}).
 While this might suggest redundancy in latter ReEigs, it's important to note that this result is specific to the dataset and post-training state.
Nonetheless, it is interesting to note that the majority of below-threshold eigenvalues were effectively "rectified" by the first BiMap layer.

TSMNet's success with $N_{BiRe}=1$ and a fixed size, made possible by its fixed output size spatial convolution layer, demonstrates an alternative viable method for constructing a \ac{drn}.

\subsection{Insights from Analyses}
\label{Disc: Analyses}

\subsubsection{Frequency Analysis}
\label{Disc: Freqs}

\begin{figure}
    \centering
    \includegraphics[width=\textwidth]{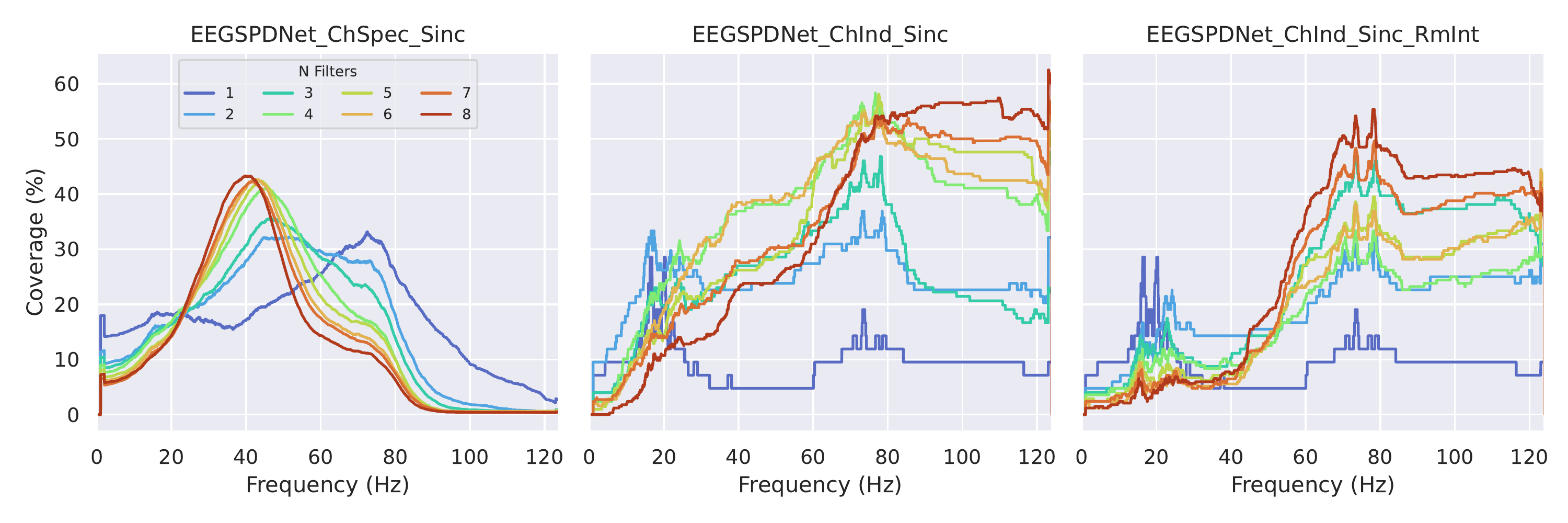}
    \caption{\color{black}
    \textbf{Frequency Band Selection Distribution for Sinc-\ac{eespdnet}.}
    A general description of the frequency band distribution plots can be found in the caption for Figure \ref{Fig:FBSPD ChosenFreqs}.
    }
    \label{Fig: Sinc Chosen Freqs}
\end{figure}

In general, where peaks are observable, the assorted frequency domain Figures \ref{Fig:EE-SPDNet Freq Gain Spectra}, \ref{Fig:EE-SPDNet Sinc Freq Gain Spectra}, \ref{Fig: FBSPD All Chosen Freqs} \& \ref{Fig: Sinc Chosen Freqs} show peaks at physiologically plausible points for the task of motor movement decoding.
Namely, peaks can be observed at 10-20Hz, 20-35Hz and 65-90Hz, (which roughly correspond to alpha, beta and high-gamma regions) which agrees with previous literature regarding motor movement in \ac{eeg} \citep{ballMovementRelatedActivity2008, pfurtschellerMotorImageryDirect2001} and suggests that \ac{eespdnet} is using known frequency bands for classification.

High gamma region utilisation varies across classifiers. 
For Conv-\ac{eespdnet}, higher relative high gamma usage correlates with higher decoding performance, with \ac{chspec} showing the highest usage, followed by \ac{chind} and \ac{chind} RmInt.

Conversely, Sinc-\ac{eespdnet} exhibits an inverse trend in high gamma usage, with \ac{chspec} using it least and the two \ac{chind} models using it most.

In the \ac{chspec} setting, high gamma usage decreases as $N_f$ increases, while the opposite occurs for \ac{chind} and \ac{chind} RmInt.
Despite this, the Sinc model sub-types follow the same decoding trend as the Conv model, with the highest accuracies achieved by the \ac{chspec} variant.

\subsubsection{Architectural layer-by-layer Analysis of EE(G)-SPDNet}
\label{Disc: LBL}
Layer-by-layer performance analysis using Riemannian and Euclidean \acp{svm} as proxy classifiers provided insights into information flow within the trained networks. 
Even though the decoding performance of the proxy classifier does not directly represent the network's classification ability, this analysis reveals some interesting patterns.

The \ac{rsvm} outperformed \ac{eespdnet} in at least one layer for nearly every case, albeit by small margins (<1\%). 
This shows the presence of task-relevant information inside the network that was not sufficiently extracted.
Notably, when $N_f=1$, \ac{rsvm} outperformed EEG-SPDNet by larger margins, with accuracies dropping most after the first BiMap-ReEig layer.
This suggests information loss in the initial transformation, possibly due to under-parameterisation (i.e. the input covariance matrix is compressed too aggressively) .
This effect is reduced as the overall increase in network parameterisation (via increased $N_F$), supporting possible under-parameterisation.

The Euclidean SVM never outperformed SPDNet but showed interesting underperformance patterns.
Channel specific Conv-\ac{eespdnet} performed best overall, requiring both channel specificity and Conv filtering.
Uniquely, its final two layers often had the worst SVM accuracies for $N_f>2$, implying a distinct data transformation resulting in higher final accuracies but lower Euclidean space accuracies.
How this final BiMap layer transformation relates to other results unique to the \ac{chspec} Conv-\ac{eespdnet} is still unclear.

\subsection{Limitations}
\label{Disc: Limitations}

The results presented in this study are subject to several limitations. 
Computational constraints restricted the extent of hyperparameter space exploration, with the initial sweep of learning rates and weight decays being relatively coarse. 
Network width exploration was limited to $N_f=8$, though performance continued to increase without plateauing for \ac{eespdnet}, suggesting potential for further improvements with wider networks. 
Furthermore, aside from an initial sweep, we did not explore network depth, changes to which could potentially yield decoding benefits.
In particular, we also did not explore any variations in the BiMap reduction factor.
Therefore the assessment of the network decoding ability should be qualified by these limitations.

Additionally, the frequencies selected by learned filterbanks does not guarantee network utilisation of these frequencies. 
As filterbank width increases, either through higher $N_f$ or \ac{chspec} filtering, attributing decoding performance to individual filters becomes more challenging.
This is also true for the multiband analysis, manual inspection of the Electrode-Frequency relevance plots (Figures \ref{Fig: CovgradP5} \& \ref{Fig: CovgradP12}) suggest their existence and potential importance for \textit{some} models/participants, but this is not guaranteed.
Furthermore, conclusions based off aggregated occurrences (Figure \ref{Fig: Multiband}) are limited  by the detection method's accuracy.

\subsection{Future}
\label{Disc: Future}
\textcolor{black}{In addition to the previously mentioned approaches such as NAS \citep{sukthankerNeuralArchitectureSearch2021}, there are a number of other areas in which the ideas presented here could be improved.}
A comprehensive exploration of additional datasets, including other BCI datasets from MOABB \citep{jayaramMOABBTrustworthyAlgorithm2018}, EMG \citep{balasubramanianEMGViableAlternative2018}, and in-ear EEG \citep{yariciEarEEGSensitivityModelling2022}, would enable better model comparison and performance verification.
Furthermore, adapting the models to other BCI paradigms such as SSVEP/P300  may reveal new aspects of model performance and expand its generalisability.

Methodological enhancements could include augmenting the covariance matrix with additional information such as done by \citet{wangCovarianceFeatureRepresentation2015}, including a temporal covariance matrix \citep{hajinorooziPredictionFatiguerelatedDriver2017}, using functional connectivity information \citep{chevallierRiemannianGeometryCombining2022}, incorporating additional Riemannian layers, such as Riemannian Batch Norm \citep{brooksRiemannianBatchNormalization2019, koblerSPDDomainspecificBatch2022} (although this could be included as part of a NAS), spatial convolution layers \citep{schirrmeisterDeepLearningConvolutional2017, koblerSPDDomainspecificBatch2022} and exploring learnable time-window selection (in a similar approach to optimised frequency band selection).

Further investigation of multiband filters is warranted and could involve refining the peak detection algorithm, developing filter relevance scores using gradient information, and creating a multiband sinc layer for SPDNet to compare with Sinc-\ac{eespdnet} and Conv-\ac{eespdnet}.

For \ac{fbspdnet}, there was no specific motivation for using a Bayesian optimiser over any other black-box optimiser, therefore exploring optimisers could yield improvements.
Additionally, performance boosts could likely be achieved by exploring the optimisation hyperparameters or alternative proxy classifiers etc.

Finally, our the hyperparameters of the networks could be explored further.
As discussed in the limitations, higher values of $N_f$ may yield further performance boosts.
Additionally, $N_BiRe$ could be further explored, especially in relation to the BiMap reduction factor.
This could even be used to increase the size of the input SPD matrix, before compressing it later in the network.
This could help capture the information that the LBL analysis suggests to have been poorly extracted in the early network.

\section{Conclusion}
\label{Conclusion}
\color{black}
In this paper\textcolor{black}{,} we have presented two novel \acfp{drn} with learnable filterbanks (\ac{eespdnet} \& \ac{fbspdnet}) for decoding EEG BCI data, which were then shown to have state-of-the-art performance on public motor movement and motor imagery dataset\textcolor{black}{s}.
\textcolor{black}{EEGSPDNet achieved the highest decoding accuracy among all proposed and comparison models, with statistically significant improvements over all but one competitor.}
These models have therefore shown improvement in what is arguably the most important factor of a \ac{bci} - raw decoding performance, while also providing a black-box approach to learning filterbank architecture.
Standard filterbank architecture design is essentially a grid-search over the frequency band space, and the learned filterbank approach of \ac{fbspdnet} \& \ac{eespdnet} allow for an optimised search of the filter space.
In particular, \ac{eespdnet}, which encapsulates the filterbank optimisation as a convolutional layer, allows for this optimisation step to be completed enclosed within a single model, in an \textit{end-to-end} manner.
Furthermore\textcolor{black}{,} the convolutional layer was shown to be flexible and consistently selecting physiologically plausible frequency regions.
Finally\textcolor{black}{,} the \ac{lbl} analysis shows there is still room for improvement on the both of the proposed models, with a number of potential avenues of further exploration.
\textcolor{black}{In summary, through our strong decoding results, our exploration of learnable filterbank methodologies and hyperparameters, as well as our analysis into resulting network behaviour, we have established a foundational basis for end-to-end \ac{drn} design and analysis.}

\section*{Data and Code Availability}
\label{Sec: Methods/DataCodeAvailability}

\textcolor{black}{All} datasets used in this study are publicly available, and were accessed via MOABB \citep{jayaramMOABBTrustworthyAlgorithm2018}.

Code for the models can be found on the public GitHub repo \url{https://github.com/dcwil/eegspdnet}.

\section*{Author Contributions}

Daniel Wilson: Conceptualization, Methodology, Software, Visualisation, Writing - Original draft preparation. Robin T. Schirrmeister: Conceptualization, Writing - Reviewing and Editing, Supervision. Lukas A. W. Gemein: Writing - Reviewing and Editing, Supervision. Tonio Ball: Conceptualization, Writing - Reviewing and Editing, Supervision, Funding acquisition.

\section*{Funding}

This work was funded by BMBF Grants: KIDELIR-16SV8864, DiaQNOS-13N16460 \& Renormalized Flows-1IS19077C and DfG grant AI-Cog-BA 4695/4-1

\section*{Declaration of Competing Interests}

The authors declare no competing interests.

\section*{Acknowledgements}

The authors acknowledge support by the state of Baden-Württemberg through bwHPC
and the Deutsche Forschungsgemeinschaft (DFG, German Research Foundation) through grant no INST 39/963-1 FUGG (bwForCluster NEMO).

\section*{Supplementary Material}

\subsection{Proof for Equation \ref{Eq:Concatenation}}

A definition for the definiteness of a matrix is for a real symmetric matrix, $M$:
\begin{equation}
    \mathcal{S}_n := \{\,M = M^\top \mid m_{ij} \in \mathbb{R}^n \}
\end{equation}
the scalar $z^\top M z$ will be positive (or non-negative in the case of \ac{psd} matrices) for every non-zero column vector, $z \in \mathbb{R}^n$.
Formally this is:

Let $M \in \mathcal{S}_n$. $M$ is said to be positive-definite if:
\begin{equation}
    z^\top M z > 0, \,\, \forall \,\, z \in \mathbb{R}^n - \{0\}
    \label{PD_definition}
\end{equation}

Let $S_n$ and $S_m$ be \ac{spd} matrices of size $n$ and $m$, respectively.
Let $Conc(S_n, S_m)$ be the mapping: $\mathcal{S}^{pd}_n \times \mathcal{S}^{pd}_m \Rightarrow \mathcal{S}^{pd}_{n + m}$ such that the output is a block diagonal matrix of the form:
\begin{equation}
    C = Conc(S_n, S_m) = 
    \begin{pmatrix}
    S_n & 0_{n \times m} \\
    0_{m \times n} & S_m \\
    \end{pmatrix}
\end{equation}

Applying Equation \ref{PD_definition} we get:
\begin{equation}
    z^\top C z =
    \begin{pmatrix}
    z_n^{\textcolor{black}{\top}} & z_m^{\textcolor{black}{\top}}
    \end{pmatrix}
    \begin{pmatrix}
    S_n & 0_{n \times m} \\
    0_{m \times n} & S_m \\
    \end{pmatrix}
    \begin{pmatrix}
    z_n \\ 
    z_m \\
    \end{pmatrix}
\end{equation}

which reduces to:

\begin{equation}
    z_n^\top S_n z_n + z_m^\top S_m z_m
\end{equation}

which will be $>0$ provided $S_n$ and $S_m$ are positive definite, which was the initial assertion.

\subsection{Frequency Gain Spectra}
\label{Sec: Appdx FreqGainSpec}

The order of calculations for frequency gain spectra (e.g. Figure \ref{Fig:EE-SPDNet Freq Gain Spectra}) is as follows:
Data for a single model is of the form:
\begin{equation}
    O = P \times S \times T \times E \times t_O
\end{equation}
where $P$ is the number of participants, $S$ is the number of initialisation seeds for each network, $T$ is the number of trials for each participant, $E$ is the number of electrodes and $t_O$ is the length of a trial in samples.

After going through the convolutional layer of a network the data array is of the form:
\begin{equation}
    C = P \times S \times T \times Ch \times t_C
\end{equation}
where $Ch$ is the number of channels ($E$ times number of \textcolor{black}{filters, $N_f$}) and $t_C$ is the length of a trial after convolution.

Then the Fourier transform of the signals was calculated:
\begin{equation}
    O \xrightarrow{FFT} \widetilde O, \,\, C \xrightarrow{FFT} \widetilde C
\end{equation}

With the length of the signals in $\widetilde O$ \& $\widetilde C$ being reduced to $\widetilde t_O$ \& $\widetilde t_C$, respectively.

Then the signals were averaged across trials:
\begin{align}
    & \widetilde O \xrightarrow{AVG_T} \textcolor{black}{\overline{\textcolor{black}{\widetilde O}}} : P \times S \times E \times \widetilde t_O \\
    & \widetilde C \xrightarrow{AVG_T} \textcolor{black}{\overline{\textcolor{black}{\widetilde C}}} : P \times S \times Ch \times \widetilde t_C
\end{align}

\textcolor{black}{$\overline{\widetilde C}$} was then interpolated (cubic) such that its signals were of the same length as \textcolor{black}{$\overline{\widetilde C}$}
\begin{equation}
    \textcolor{black}{\overline{\widetilde C}} \xrightarrow{Interp} \\textcolor{black}{overline{\widetilde C}}: P \times S \times Ch \times \widetilde t_O
\end{equation}

The electrode arrays in \textcolor{black}{$\overline{\widetilde O}$} were then duplicated such that they matched the number of channels in \textcolor{black}{$\overline{\widetilde C}$}.
\begin{equation}
    \textcolor{black}{\overline{\widetilde O}}: P \times S \times Ch \times \widetilde t_O
\end{equation}

The signal gain (in dB), $G$, was then calculated:
\begin{equation}
    G = 20 log_{10}\left(\frac{\textcolor{black}{\overline{\widetilde O}}}{\textcolor{black}{\overline{\widetilde C}}}\right)
\end{equation}

\subsection{Supplementary Comparisons}

\begin{figure}
    \centering
    \includegraphics[width=\textwidth]{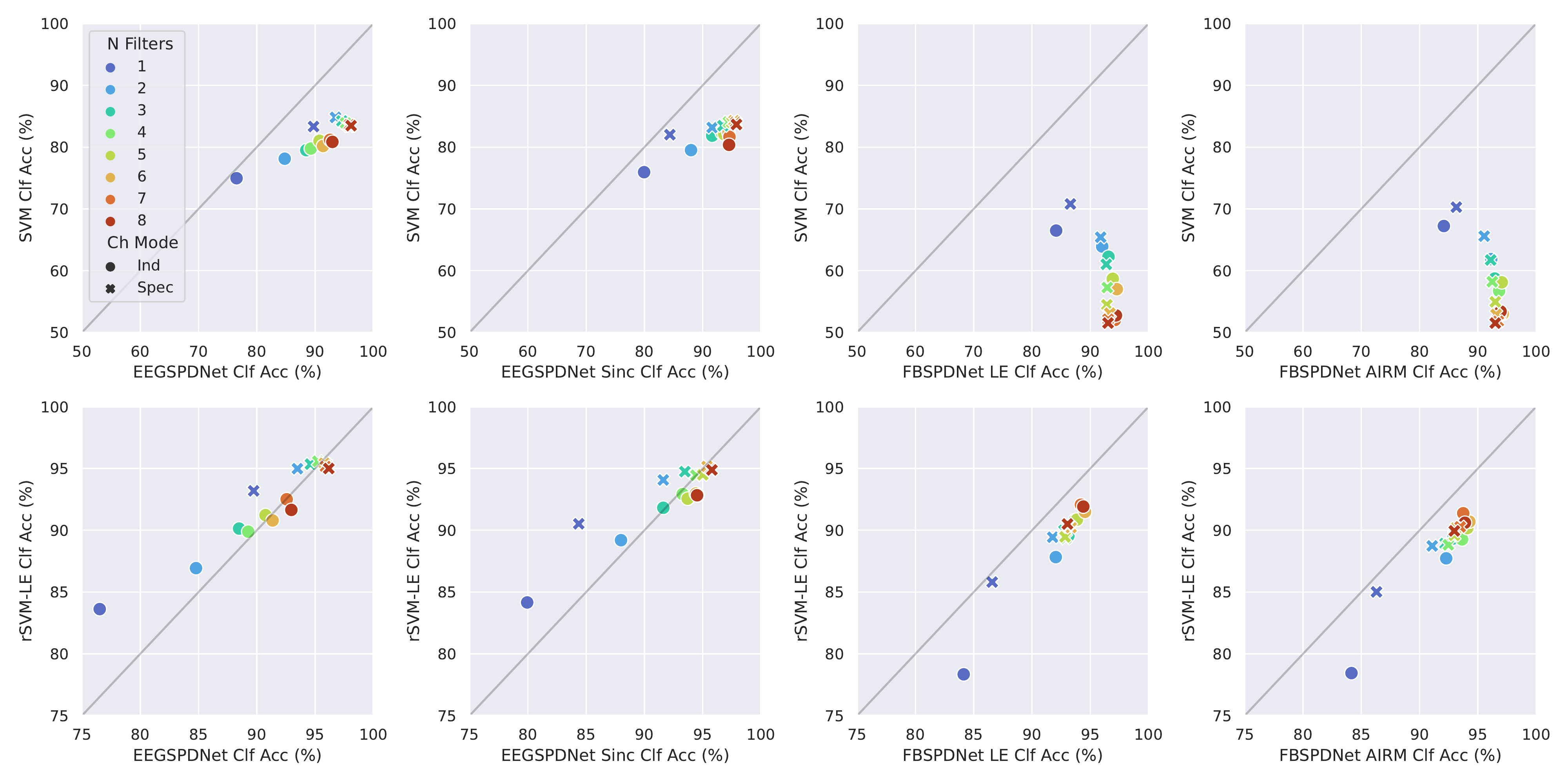}
    \caption{
    \textcolor{black}{
    \textbf{SVM and rSVM-LE Accuracies on Post-Filterbank Covariance Matrices.}
    Y-axes show the SVM (top row) or rSVM (bottom row) classification accuracies.
    X axes show the \ac{eespdnet} or \ac{fbspdnet} classification accuracies.
    Trials were filtered with the learned filterbank of the associated \ac{eespdnet} or \ac{fbspdnet}.
    Note the different axes limits between the top and bottom rows.
    Hue and marker type respectively highlight $N_f$ and channel specificity.
    }}
    \label{Fig: SVM Scatterplots}
\end{figure}
\subsection{Additional LBL Figures}
\label{Sec: Appdx LBL}

\subsection{Additional Chosen Frequency Distributions}
\label{Sec: Appdx: Chosen Freqs}

\subsection{Software}
\label{Appdx: Software}
All code was written with Python 3.8 and featured the following libraries:
\begin{itemize}
    \item SciPy \citep{virtanenSciPyFundamentalAlgorithms2020}
    \item NumPy \citep{harrisArrayProgrammingNumPy2020}
    \item MNE \citep{gramfortMEGEEGData2013}
    \item PyTorch \citep{paszkePyTorchImperativeStyle2019}
    \item Pandas \citep{mckinneyDataStructuresStatistical2010}
    \item Sci-Kit Learn \citep{pedregosaScikitlearnMachineLearning2011}
    \item Braindecode \citep{schirrmeisterDeepLearningConvolutional2017}
    \item Seaborn \citep{waskomSeabornStatisticalData2021}
    \item Matplotlib \citep{hunterMatplotlib2DGraphics2007}
    \item \textcolor{black}{DabEst} \citep{hoMovingValuesData2019}
    \item \textcolor{black}{bayes\_opt} \citep{bayesopt_package}
    \item \textcolor{black}{geoopt} \citep{kochurovGeooptRiemannianOptimization2020}
\end{itemize}

Furthermore, we used and adapted code from the following libraries
\begin{itemize}
    \item SPDNet (Python): https://github.com/adavoudi/spdnet
    \item TorchSPDNet (Python): https://gitlab.lip6.fr/schwander/torchspdnet/-/tree/master/
    \item SPDNet (Matlab): https://github.com/zhiwu-huang/SPDNet
    \item \textcolor{black}{TSMNet (Python): https://github.com/rkobler/TSMNet}
\end{itemize}

Upon publication a GitHub repository containing the code used for generating the results will be made public.

\subsection{Hardware}

All computations were performed using bwForCluster NEMO, hardware details of which can be found here: \url{https://wiki.bwhpc.de/e/NEMO/Hardware}

\subsection{Datasets}
\label{Appdx:Data}

\begin{table}
\begin{threeparttable}
\caption{\textcolor{black}{
\textbf{Dataset Information.}
Table showing dataset metadata, more details can be found in the papers associated with each dataset (see Table),  on the moabb website (\url{http://moabb.neurotechx.com/docs/dataset_summary.html}) and in the repository associated with this study (see Section \ref{Sec: Methods/DataCodeAvailability}).
}
\label{Tab: Datasets}}
{\color{black}\begin{tabular}{@{}llllll}
\toprule
& BNCI2014001\tnote{$a$} & BNCI2014004\tnote{$b$} & Lee2019 MI\tnote{$c$}  & Schirrmeister2017\tnote{$d$}  & Shin2017A\tnote{$e$} \\
\midrule
     \# Electrodes & 22 & 3 & 62 & 44 & 30 \\
     \# Total Trials & 5184 & 6520 & 11000 & 13440 & 1740 \\
     \# Participants & 9 & 9 & 54 & 14 & 29 \\
     \# Classes & 4 & 2 & 2 & 4 & 2 \\
     Samp Freq & 250 & 250 & 250 & 250 & 200 \\
     Prepro Filtering & (4, 38) & (4, 38) & (4, 38) & (4, 124) & (4, 38) \\
     Batch Size & 120 & 120 & 80 & 256 & 20 \\
\bottomrule
\end{tabular}}
\begin{tablenotes}
\item[${a}$] \citep{tangermannReviewBCICompetition2012}
\item[${b}$] \citep{leebBrainComputerCommunication2007}
\item[${c}$] \citep{leeEEGDatasetOpenBMI2019}
\item[${d}$] \citep{schirrmeisterDeepLearningConvolutional2017}
\item[${e}$] \citep{shinOpenAccessDataset2017} 
\end{tablenotes}
\end{threeparttable}
\end{table}

\subsection{Parameter Sweeps}
\label{Sec: Appdx ParamSweeps}

\subsection{Model \textcolor{black}{Hyper}parameters}
In this section the \textcolor{black}{hyper}parameters for the models can be found.

\subsubsection{General}
\begin{itemize}
    \item Number of random seeds: 3
    \item SPD Estimator: Sample Covariance Matrix
\end{itemize}
\subsubsection{SPDNet}
\begin{itemize}
    \item LR Scheduler: Cosine Annealing
    \item Epochs: 1000
    \item Loss: Cross-entropy loss
    \item \textcolor{black}{ReEig Threshold: 5e-4}
\end{itemize}
\subsubsection{EE(G)-SPDNet}
\begin{itemize}
    \item Convolutional Kernel Length: 25
\end{itemize}
\subsubsection{\textcolor{black}{FB}SPDNet}
\textcolor{black}{SVM parameters were chosen heuristically during initial testing.}
\begin{itemize}
    \item BO Stopping Criteria: 1000 iterations OR 12 hours
    \item SVM C: 1
    \item SVM Kernel: linear
    \item Cross-validation: Stratified 3-fold
\end{itemize}
\color{black}
\subsection{Peak Detection}
We used the "scipy.signal.find\_peaks" function, with the following heuristically determined parameters.
Other parameters were left at their default value.
\begin{itemize}
    \item \textcolor{black}{Peak Width: roughly 1Hz}
    \item \textcolor{black}{Peak Height: 1.5x Std Dev amplitude}
\end{itemize}
\color{black}
\printbibliography

\appendix



\end{document}